\pdfoutput=1

\documentclass[11pt]{article}

\usepackage[preprint]{acl}

\usepackage{times}
\usepackage{latexsym}

\usepackage[T1]{fontenc}

\usepackage[utf8]{inputenc}

\usepackage{microtype}

\usepackage{inconsolata}

\usepackage{graphicx}

\usepackage[utf8]{inputenc} 
\usepackage[T1]{fontenc}    
\usepackage{hyperref}       
\usepackage{url}            
\usepackage{booktabs}       
\usepackage{amsfonts}       
\usepackage{nicefrac}       
\usepackage{microtype}      
\usepackage{algorithm}
\usepackage{algpseudocodex}
\usepackage{amsmath}  
\usepackage{graphicx} 
\usepackage{anyfontsize}

\usepackage{bbding}
\usepackage{wrapfig}
\usepackage{upquote}
\usepackage{subcaption}
\usepackage{multirow}

\usepackage{textcomp}  
\usepackage{scalerel}  
\usepackage{xspace}
\usepackage{arydshln}
\usepackage{listings}
\usepackage{fancyvrb}
\usepackage{titletoc}
\usepackage{amssymb}
\usepackage{pifont}

\usepackage{colortbl}

\usepackage[breakable]{tcolorbox}
\tcbuselibrary{hooks, breakable, skins}
\usepackage{lipsum}

\definecolor{darkgreen}{RGB}{50,100,0}
\definecolor{darkred}{RGB}{200, 0, 0}
\newcommand{\cmark}{\textcolor{darkgreen}{\ding{51}}} %
\newcommand{\xmark}{\textcolor{darkred}{\ding{55}}}

\definecolor{wkyellow}{RGB}{255,241,177}
\definecolor{lightblue}{HTML}{CCE5FF}
\definecolor{lightgray}{gray}{0.9}
\definecolor{goodblue}{HTML}{0071bc}

\usepackage{soul} 
\definecolor{error}{rgb}{1,0.8,0.8} 
\DeclareRobustCommand{\hlerror}[1]{{\sethlcolor{error}\hl{#1}}}

\definecolor{lightgray}{gray}{0.9}

\lstdefinestyle{casestyle}{
  basicstyle=\footnotesize\ttfamily,
  columns=fullflexible,
  breaklines=true,
  breakindent=0pt,     
  frame=lines,
  extendedchars=true,
  escapechar=@,
  literate={á}{{\'a}}1 {ã}{{\~a}}1 {é}{{\'e}}1 {£}{{\pounds}}1 {–}{{-}}1 {’}{{'}}1,
}

\title{Markov Chain of Thought for Efficient Mathematical Reasoning}

\author{%
  Wen Yang\thanks{\ \ Equal contribution}, Minpeng Liao\footnotemark[1]\thanks{\ \ Corresponding author}, Kai Fan\footnotemark[1]\footnotemark[2] \\
  Alibaba Tongyi Lab \\
  \texttt{\{hechu.yw, minpeng.lmp, k.fan\}@alibaba-inc.com}
}

\begin{document}

\maketitle

\begin{abstract}
Chain of Thought (CoT) of multi-step benefits from the logical structure of the reasoning steps and task-specific actions, significantly enhancing the mathematical reasoning capabilities of large language models. 
As the prevalence of long CoT, the number of reasoning steps exceeds manageable token limits and leads to higher computational demands. 
Inspired by the fundamental logic of human cognition, ``derive, then reduce'', we conceptualize the standard multi-step CoT as a novel Markov Chain of Thought (MCoT). 
In this study, we consider the mathematical reasoning task, defining each reasoning step as text accompanied by a Python code snippet. 
To facilitate a longer reasoning path, self-correction is enabled through interactions with the code interpreter. 
Our MCoT aims to compress previous reasoning steps into a simplified question, enabling efficient next-step inference without relying on a lengthy KV cache. 
In our experiments, we curate the \texttt{MCoTInstruct} dataset, and the empirical results indicate that MCoT not only significantly enhances efficiency but also maintains comparable accuracy. 
While much remains to be explored, this work paves the way for exploring the long CoT reasoning abilities of LLMs.
The code is available at \url{https://github.com/james-yw/Markov-Chain-of-Thought}.
  
\end{abstract}

\section{Introduction}
\label{sec:intro}


\begin{figure}[!ht]
\centering

\includegraphics[width=0.9\linewidth]{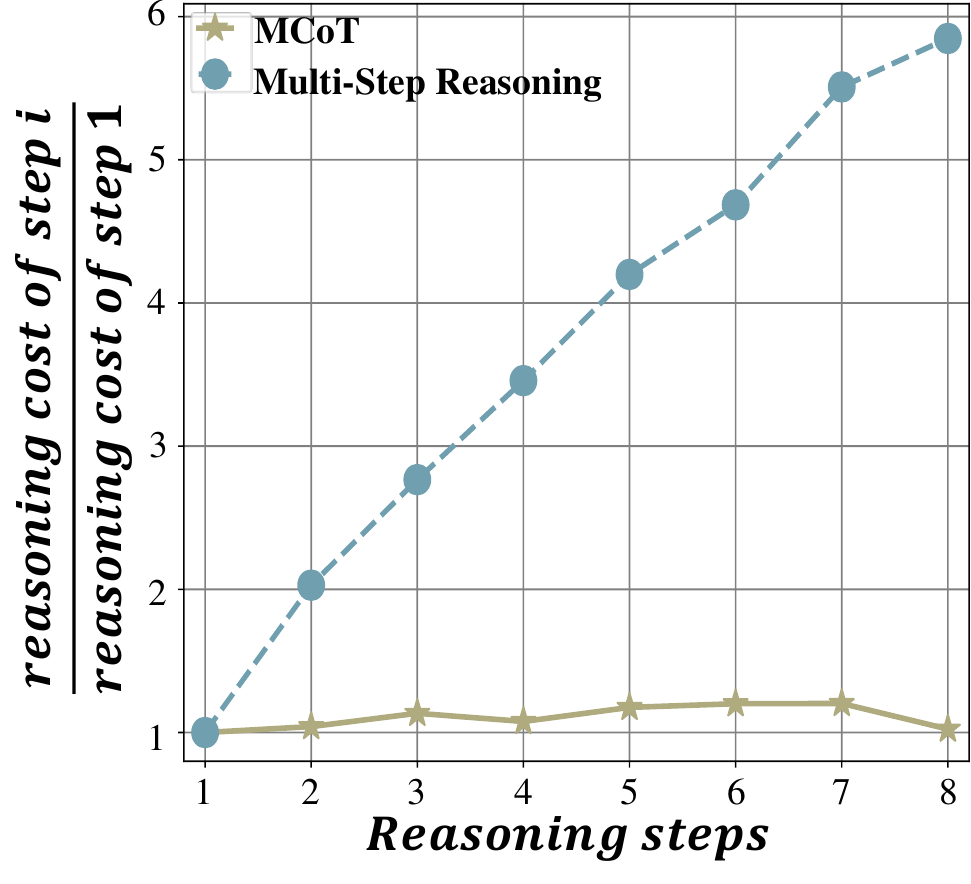} 
\label{fig:intro_sub_eff}
\caption{Comparison of reasoning efficiency between MCoT and Multi-Step Reasoning (MSR), showing the variation in reasoning time costs for both methods relative to \textit{step 1} as the number of reasoning steps increases.}
\vspace{-6mm}
\label{fig:intro_mcot}
\end{figure}


\begin{figure*}[!ht]
\centering
\includegraphics[width=1.0\textwidth]{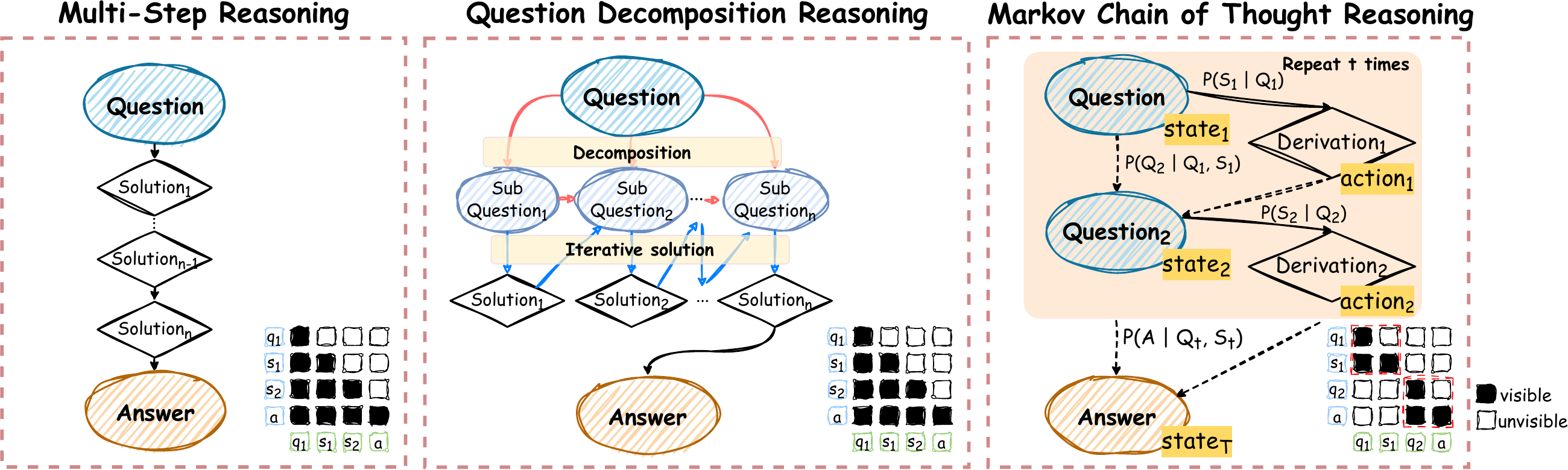}
\caption{
\textbf{Schematic illustrating various approaches to mathematical reasoning with LLMs and their reasoning efficiency.} 
The masked demonstrations across different approaches show that the efficiency of MCoT is similar to that of the blockwise masking approach, while the efficiency of MSR and question decomposition reasoning is more akin to that of the vanilla masking.
}
\label{fig:intro_cmp}
\vspace{-4mm}
\end{figure*}


With the rapid advancement of large language models (LLMs), these models have demonstrated remarkable progress in a wide range of language tasks~\citep{brown2020language,ouyang2022training,taori2023stanford,yang2021fudge}. 
However, they still face significant challenges when engaging in complex and symbolic reasoning tasks, particularly in mathematical reasoning~\citep{cobbe2021training, hendrycks2021measuring}. 

Most existing works have sought to enhance the mathematical reasoning capabilities of LLMs.
These efforts can be broadly categorized into two approaches: single-step reasoning~\citep{yu2023metamath, yue2023mammoth} and multi-step reasoning~\citep{gou2023tora, wang2023mathcoder,liao2024mario, lu2024mathgenie,chen2024alphamath,chen2024step}.
Single-step reasoning completes the reasoning task through one inferential step; Multi-step reasoning, which utilizes the reasoning traces (CoT) and task-specific actions (Code Interpreter), has been empirically demonstrated to boost the complex reasoning abilities of LLMs. 
Nevertheless, as the number of reasoning steps increases,~\citet{levy2024same} finds that multi-step reasoning may become susceptible to accumulative errors or hallucinations. 
Furthermore, multi-step reasoning involving processing long CoT demands greater computational resources, which renders the inference process practically inefficient.


To address the inefficiency inherent in multi-step reasoning, our approach is deeply influenced by understanding how humans navigate complex reasoning tasks. 
Research on human cognition~\citep{simon1971human,polya2004solve,meadows2008thinking} highlights two pivotal phases: \textit{derivation} and \textit{reduction}. 
When tackling a complex problem, the process often starts with deriving intermediate variables or partial solutions, rather than seeking an immediate full solution. 
These intermediate steps help clarify key aspects of the problem, which can then be simplified through reduction.
The process of derivation and reduction is not linear but rather \textit{iterative}.
After each reduction, the newly simplified problem becomes an \textit{independent} question, prompting further exploration for solutions. 
These new solutions could facilitate further reductions, gradually simplifying the problem and moving it toward the final answer.
To formalize this reasoning process, we utilize the Markov chain to model the entire sequence of reasoning. 
This innovative approach is termed Markov Chain of Thought (MCoT). 

The Markovian property has been explored in the mathematical proof of formal languages (FL), such as  Lean~\cite{moura2021lean}.
For example, executing a Lean tactic—equivalent to a reasoning step—transforms an original hypothesis (which may be none) into a new one while discarding the original. 
We aim to extend this idea to natural language (NL) for mathematical reasoning.
Figure~\ref{fig:intro_cmp} illustrates the principle of MCoT approach.
In MCoT, we initially frame the question as a particular \texttt{state} and consider the first derivation step as an associated \texttt{action}, then envision the complete solution as a sequential series of transitions between states. 
In contrast to question decomposition approaches~\citep{zhou2022least, dua2022successive, huang2023question, radhakrishnan2023question}, the Markov property inherent in MCoT framework guarantees that any \texttt{state} (question) can directly lead to the final answer. 
However, question decomposition breaks down a problem into multiple sub-questions, each yielding only a partial answer. It is the cumulative aggregation of these partial answers that ultimately generates the final answer.
Unlike multi-step reasoning, MCoT clears the KV cache of the context after simplifying questions, enabling it to support longer CoT.
Therefore, MCoT offers the notable advantage that the reasoning's time or memory demands do not linearly or quadratically increase with the number of reasoning steps. 

Empirically, we have developed a dataset specifically designed for MCoT reasoning, called \texttt{MCoTInstruct}. In order to maximize the utilization of existing data resources, we provide a reproducible pipeline starting from the multi-step reasoning datasets to construct the MCoT dataset. 
This dataset, originating from GSM8K \citep{cobbe2021training} and MATH \citep{hendrycks2021measuring}, has been further refined through a combination of GPT-4 annotations and self-distillation processes.

In summary, Our contributions are as follows:\\
\noindent
\textbf{1.} We propose an innovative framework Markov Chains of Thought (MCoT), by exploiting the Markov property and envisioning the solution process as a series of transitions between states.\\
\noindent
\textbf{2.} We constructed a \texttt{MCoTInstruct} dataset on mathematical reasoning task to facilitate research community.\\
\noindent
\textbf{3.} Extensive experiments demonstrate that, with a maximum number of eight steps, MCoT achieves an average reasoning efficiency that is $1.90\times$ faster than traditional multi-step reasoning and maintains superior to multi-step reasoning  accuracy.\\
\noindent
\textbf{4.} As MCoT provides a new pathway for exploring advanced reasoning abilities, we will release our dataset and code to facilitate further research and development within the community. 

\section{Markov Chain of Thought Enable Mathematical Reasoning}

\begin{algorithm}[t]
\small
\caption{\small{Construction of the MCoT Seed Dataset}}\label{alg:seed data}
\begin{algorithmic}[1]
\State $\mathcal{D}_{\text{origin}} \gets$ Load initial GSM8K and MATH dataset
\State Define data format in $\mathcal{D}_{\text{origin}}$: $\tau_1 = (\mathbf{q}_1, \mathbf{s}_{1:T}, \mathbf{a})$, where $T\geq 1$.
\State Train model $M_{\text{verify}}$ using $\mathcal{D}_{\text{origin}}$
\State $\mathcal{D}_{\text{seed}} = [~]$
\State $\tilde{\mathcal{D}}_{\text{origin}} = [~]$
\For{each instance in $\mathcal{D}_{\text{origin}}$}
    \If{$T = 1$}
        \State Add $(\mathbf{q}_1, \mathbf{s}_1, \mathbf{a})$ to $\mathcal{D}_{\text{seed}}$
    \Else
        \State Use GPT-4 to generate reduced question $\mathbf{q}_2$ from $(\mathbf{q}_1, \mathbf{s}_1)$
        \State Inference mode: $M_{\text{verify}}(\mathbf{q}_2)$ to obtain $\left( \mathbf{s}^\prime_{1:T^{\prime}}, \mathbf{a}^\prime \right)$
        \If{$\mathbf{a}^\prime = \mathbf{a}$}                                 \Comment{Independence Test}
            \State Add $(\mathbf{q}_1, \mathbf{s}_1, \mathbf{q}_2)$ to $\mathcal{D}_{\text{seed}}$
            \State Add new multi-step reasoning path $\tau_2 = (\mathbf{q}_2, \mathbf{s}^\prime_{1:T^{\prime}}, \mathbf{a})$ to $\tilde{\mathcal{D}}_{\text{origin}}$
        \EndIf
    \EndIf
\EndFor
\State $\mathcal{D}_{\text{origin}} \leftarrow \tilde{\mathcal{D}}_{\text{origin}}$ then go to Line 5
\end{algorithmic}
\end{algorithm}

\subsection{Markov Chain of Thought Reasoning}

For mathematical problem $\mathbf{q}$, we assume that with each successful derivation step, the original problem can be incrementally simplified into a series of less complex problems, eventually leading to the final answer $\mathbf{a}$. 
Concretely, if we denote the original problem by $\mathbf{q}_1$ and the first derivation step as $\mathbf{s}_1$, we can define the generation of new problems as $p(\mathbf{q}_t | \mathbf{q}_{t-1}, \mathbf{s}_{t-1})$. 
Consequently, the subsequent derivation step relies entirely on the newly formulated question. 
This indicates that the process adheres to the Markov property, which implies memorylessness (\emph{i.e.}, the future state depends only on the current state and not on the sequence of events that preceded it).
\begin{equation}\label{eq:markov}
    p(\mathbf{s}_t|\mathbf{q}_{t' \leq t}, \mathbf{s}_{t' < t}) = p(\mathbf{s}_t | \mathbf{q}_t) 
\end{equation}
For a question with $T$ derivation steps, we are interested in maximizing the log-likelihood of the joint distribution of all steps. 
With the above assumption, we have the following objective.
\begin{equation}
\begin{aligned}
    \mathcal{L} &= \log p(\mathbf{a}, \mathbf{s}_{1:T}|\mathbf{q}_1) \\
    &= \log \Big( p(\mathbf{s}_1 | \mathbf{q}_1) 
    \, \mathbb{E}_{\mathbf{q}_{2:T}} \Big[ p(\mathbf{a}|\mathbf{q}_{T}, \mathbf{s}_{T}) \\
    &\quad \times \prod_{t=2}^{T} p(\mathbf{s}_t|\mathbf{q}_t) \, p(\mathbf{q}_t | \mathbf{q}_{t-1}, \mathbf{s}_{t-1}) \Big] \Big)
\end{aligned}
\end{equation}
However, this objective is intractable due to the requirement of integrating latent variables $\mathbf{q}_{2:T}$. 
As a surrogate, we turn to Monte Carlo integration, employing sampling techniques for feasibility. 
When we set the sampling size to 1, and if $\tilde{\mathbf{q}}_{2:T}$ represents the sequence of sampled reduction questions of intermediate derivation steps, the objective $\mathcal{L}$ can be approximated as follows. 
\begin{equation}
\small
\begin{aligned}\label{eq:loss}
    &\log \left( p(\mathbf{s}_1|\mathbf{q}_1) p(\mathbf{a}|\tilde{\mathbf{q}}_{T}, \mathbf{s}_{T}) \prod_{t=2}^{T} p(\mathbf{s}_t|\tilde{\mathbf{q}_t}) p(\tilde{\mathbf{q}}_t|\mathbf{q}_{t-1}, \mathbf{s}_{t-1}) \right)\\
    &\text{Denote $\mathbf{q}_1 = \tilde{\mathbf{q}}_1$}, \\
    &\quad= \log p(\mathbf{a}|\tilde{\mathbf{q}}_{T}, \mathbf{s}_{T}) + \sum_{t=1}^T \log  p(\mathbf{s}_t|\tilde{\mathbf{q}_t}) \\
    &\hspace{28mm} + \sum_{t=1}^{T-1} \log p(\tilde{\mathbf{q}}_{t+1}|\mathbf{q}_t, \mathbf{s}_t) \\
    &\quad = \log p(\mathbf{s}_T, \mathbf{a}|\tilde{\mathbf{q}}_T) + \sum_{t=1}^{T-1} \log p(\mathbf{s}_t, \tilde{\mathbf{q}}_{t+1}|\tilde{\mathbf{q}}_t) \\
    &\text{Denote $\mathbf{a}=\tilde{\mathbf{q}}_{T+1}$}, \\
    &\quad=\sum_{t=1}^{T} \log p(\mathbf{s}_t, \tilde{\mathbf{q}}_{t+1}|\tilde{\mathbf{q}}_t)
\end{aligned}
\end{equation}
In Eq.~(\ref{eq:loss}), the first equation  unfolds the approximated loss into the summation of $2T$ independent log-likelihoods. 
The second equation applies expression re-organization with the rule of conditional probability $p(\mathbf{s}_t|\mathbf{q}_t)p(\mathbf{q}_{t+1}|\mathbf{q}_t, \mathbf{s}_t) = p(\mathbf{s}_t, \mathbf{q}_{t+1}|\mathbf{q}_t)$, resulted in $T$ new independent log-likelihoods, signifying that these components can be optimized independently. 
The third equation is to rewrite the loss into a more concise representation. 
To sum up, if \textbf{multi-step reasoning} can be transformed into \textbf{multiple independent single-step reasoning}, the training and inference become very efficient. 
This is also the core intuition to build our dataset.

\subsection{MCoTInstruct Dataset}

Our \texttt{MCoTInstruct} dataset is comprised of two components: the seed data, denoted as $\mathcal{D}_{\text{seed}}$, and the augmented self-distillation data, referred to as $\mathcal{D}_{\text{self}}$. 
To construct our dataset, we start from an available multi-step reasoning dataset, denoted as $\mathcal{D}_{\text{origin}}$, which includes the multi-step solutions from GSM8k and MATH datasets that have been further refined through GPT-4 annotations, \emph{e.g.}, 
MathCodeInstruct~\cite{wang2023mathcoder}. 
The data format of $\mathcal{D}_{\text{origin}}$ is represented as $\tau_1=(\mathbf{q}_1, \mathbf{s}_{1:T}, \mathbf{a})$, where $\mathbf{a}$ stands for the final answer, and $\mathbf{s}_t$ signifies the intermediate derivation step at time $t$. 
Particularly, we assume that $T \geq 1$, implying the solution includes at least one derivation step. 
Furthermore, this derivation step adheres to the REACT~\cite{yao2022react} style, with customized \texttt{<Text, Code, Observation>} format that integrates text analysis with executable code blocks within the process of crafting a response, effectively enhancing the precision of reasoning.

\paragraph{Seed data} To fully leverage available data resources, we have established a reproducible pipeline that iteratively extracts the required seed training instance and updates the multi-step reasoning dataset $\mathcal{D}_{\text{origin}}$. 
Algorithm \ref{alg:seed data} presents the overall pipeline designed to generate seed dataset.

First, we train a model based on DeepSeekMath 7B base model, denoted as $M_{\text{verify}}$, using the initially original dataset $\mathcal{D}_{\text{origin}}$. 
This model serves dual purposes: generation and verification. 
In the generation phase, $M_{\text{verify}}$ produces multiple multi-step solution samples for a given reduction question. 
During verification, if any final answer within these sampled solutions aligns with the answer to the original question, the corresponding reduction question is deemed acceptable.

Then, given an instance $(\mathbf{q}_1, \mathbf{s}_{1:T}, \mathbf{a})$ in $\mathcal{D}_{\text{origin}}$, if $T = 1$, we directly incorporate this triplet $(\mathbf{q}_1, \mathbf{s}_1, \mathbf{a})$ into our seed dataset. 
For other cases, we employ \texttt{GPT-4-1106-preview}~\citep{achiam2023gpt} to produce the reduction question $\mathbf{q}_2$ from $(\mathbf{q}_1, \mathbf{s}_1)$. The details of GPT-4 prompt can be found in Appendix~\ref{appendix:prompt}.
We employ $M_{\text{verify}}$ to assess whether $\mathbf{q}_2$ can yield the correct answer. 
If the outcome is accurate, it demonstrates that the reduction question is independent and does not rely on information from the previous questions and derivation steps, satisfying the required Markovian property. 
In this case, we include triplet $(\mathbf{q}_1, \mathbf{s}_1, \mathbf{q}_2)$ in the seed dataset, and a new multi-step reasoning pathway is generated as $\tau_2=(\mathbf{q}_2, \mathbf{s}^{\prime}_{1:T^\prime}, \mathbf{a})$, which is then updated to the original dataset for the next round construction. 
This implies that we continuously iterate through the process to generate triplet data until we no longer obtain any multi-step reasoning solutions with a length exceeding two. 
Unlike previous works~\cite{yue2023mammoth,gou2023tora,wang2023mathcoder,liao2024mario}, 
our approach relies solely on GPT-4 for generating a new question, typically a single sentence, instead of crafting a complete solution that includes text analysis and code snippets. 
Consequently, this method incurs significantly lower additional costs.


\paragraph{Self-distillation}
We fine-tune the DeepSeekMath 7B base model on the above seed data to obtain the \texttt{MCoTModel-initial}. 
Recognizing the limited scope of our seed data on the MATH dataset, we have adopted a self-distillation approach to substantially enhance the coverage and diversity of the dataset. We employ \texttt{MCoTModel-initial} due to its capability to generate MCoT paths, achieving accuracy rates of $77.10\%$ and $53.48\%$ on the GSM8K and MATH datasets, respectively. Given \texttt{<Question, Answer>} pairs from the training sets of GSM8K and MATH, the initial model can generate Markov reasoning paths and obtain answers. We will verify the answers to form a self-distillation dataset $\mathcal{D}_{\text{self}}$. Furthermore, by utilizing \texttt{MCoTModel-initial} with \texttt{<Question, Answer>} pairs from any dataset, we can create a self-distillation dataset.
In the Appendix~\ref{appendix:self-distillation}, we provide a detailed analysis of the impact of self-distillation.

Combining seed data $\mathcal{D}_{\text{seed}}$  and self-distilled data $\mathcal{D}_{\text{self}}$, we remove duplicate entries to form the \texttt{MCoTInstruct} dataset, which is denoted as $\mathcal{D} = \text{filter}(\{\mathcal{D}_{\text{seed}}, \mathcal{D}_{\text{self}}\})$. 
The dataset comprises 82$k$ Markov chains, totaling around 160$k$ entries, data format is shown in Appendix~\ref{appendix:data_format}. 
Uniquely, in contrast to prior studies, each training instance corresponds to the step level rather than the solution level that may include multiple steps, while Table~\ref{tab:data_compare} in the Appendix~\ref{appendix:data} compares our dataset with recently proposed mathematical reasoning datasets.

\section{Discussion}

\textbf{The insight of MCoT} is Markov chain, which frames the question as a particular \texttt{state} and considers the derivation step as an associated \texttt{action}. 
Unlike Multi-Step Reasoning (MSR), the MCoT does not depend on historical derivation steps; Furthermore, while question decomposition reasoning requires aggregating partial answers from each sub-question, any reduction question within MCoT can directly yield the final answer. 

\textbf{The core of MCoT} lies in the Markov property, which implies memorylessness, the future state depends only on the current state, not on the sequence of events that preceded it.
On one hand, we utilize Markov property to ensure that any state (or question) within a Markov Chain of Thought can directly lead to the final answer, significantly boosting efficiency within MCoT. 
On the other hand, when an error arises in an intermediate state, the memoryless feature of MCoT might propagate it, leading to errors in subsequent states as well. 
In contrast, traditional MSR is capable of accessing all previous historical contexts, potentially allowing it to correct intermediate errors. 
However, in our experiments, we empirically observed that MSR does not necessarily correct more intermediate errors than our proposed MCoT approach.

\textbf{The motivation of MCoT}, which introduces the Markov property into natural language mathematical reasoning, is inspired by the successful application of the Markov property in formal language systems, such as the \texttt{lean} programming language~\citep{moura2021lean}. In formal language systems, a statement consists of conditions and a conclusion, e.g., $h_1, h_2, h_3 \rightarrow c$. The proof process involves stepwise reductions, where each step simplifies the statement by eliminating old conditions and adding new ones, e.g., $h_3, h_4 \rightarrow c$. This allows us to focus only on the updated statement, ensuring that if the new statement is proven or solved, the original one is verified as well.

While the Markov property is theoretically sound in formal language mathematical reasoning, formal languages like \texttt{lean} are not easily accessible to non-experts. Therefore, we explore applying Markov property to natural language mathematical reasoning, which is more user-friendly. Our future work will also investigate combining natural language and formal language to balance rigor and readability.

\section{Experiments}
\subsection{Experimental Setup}

\paragraph{Implementation Details} 
We fine-tune DeepSeekMath-Base~\citep{shao2024deepseekmath}, LLemma~\citep{azerbayev2023llemma}, and LLama-3~\cite{llamma3blog} series (ranging from 7B to 70B) on the \texttt{MCoTInstruct} to evaluate the efficacy and accuracy of our MCoT framework. The implementation details are described in Appendix~\ref{appendix:param_details}.

\paragraph{Datasets} We have selected diversity evaluation datasets, encompassing both in-domain and out-of-domain datasets from various mathematical fields, to assess the models’ capabilities in mathematical reasoning. 
For the in-domain test sets, we choose GSM8K~\citep{cobbe2021training} and MATH~\citep{hendrycks2021measuring}. 
For the out-of-domain test sets, we choose the open-source OCWCourses~\citep{lewkowycz2022solving} dataset and GaoKao2023-Math-En~\citep{liao2024mario} dataset. A comprehensive overview of these datasets is presented in Appendix~\ref{appendix:test_data}.

\paragraph{Metrics} We evaluate from two perspectives: reasoning \textbf{efficiency} and reasoning \textbf{accuracy}. 
To ensure fairness when evaluating reasoning efficiency, 
we measure the average amount of cached GPU memory used per sample during inference and design a metric $E$ to measure the decoding time of each token in the average step. 
\begin{equation}\label{eq:decoding_time}
\small
    E = \frac{1}{T} \sum_{t=1}^{T} \frac{\text{time}_t}{\#\{\mathbf{s}_t\}}
\end{equation}
where $T$ is the number of steps, $\text{time}_t$ is the decoding time for step $t$, and $\#\{\mathbf{s}_t\}$ is the number of tokens at step $t$. 
For assessing reasoning accuracy, we utilize the mathematics evaluation toolkit in ~\citet{zhang2024mario}.

\paragraph{Baselines} 
We compared proprietary, open-source, and math-specific models fine-tuned on mathematical reasoning datasets. Details are in Appendix~\ref{appendix:baselines}.

\subsection{Main Results}

\begin{table*}[!ht]
    \centering
    \small
    \resizebox{0.95\linewidth}{!}{%
    \begin{tabular}{lcccccccc}
      \toprule[1.2pt] 
      Model & Size & Data & Tool & Zero & \multicolumn{2}{c}{In-domain} & \multicolumn{2}{c}{Out-of-domain} \\
      \cline{6-9}
       &  & Size & & Shot & GSM8K & MATH & OCW & GK2023$^*$ \\
      \midrule[0.8pt]
      \multicolumn{9}{c}{Proprietary Models} \\
      \midrule[0.8pt]
      GPT-4 & - & - & \xmark & \xmark & 92.0 & 42.5 & - & - \\
      GPT-4-Code & - & - & \cmark & \xmark & 92.9 & 69.7 & 30.1 & 43.6 \\
      ChatGPT & - & - & \xmark & \xmark & 80.8 & 35.5 & - & - \\
      ChatGPT(PAL) & - & - & \cmark & \xmark & 78.6 & 38.7 & - & - \\
      PaLM-2 & 540B & - & \xmark & \xmark & 80.7 & 34.3 & - & -\\
      \midrule[0.8pt]
      \multicolumn{9}{c}{Open-Source Models} \\
      \midrule[0.8pt]
      Llama-2 & 7B & - & \xmark & \xmark & 13.3 & 4.1 & 3.7& -\\
      Llama-2 SFT & 7B & - & \xmark & \cmark & 41.3 & 7.2 & - & -\\
      Llama-2 RFT & 7B & - & \xmark & \cmark & 51.2 & - & - & -\\
      Llemma & 7B & - & \xmark & \xmark & 36.4 & 18.0 & 7.7& -\\
      Llemma(PAL) & 7B & - & \cmark & \xmark & 40.1 & 21.5 & - & -\\
      CodeLlama & 7B & - & \xmark & \xmark & 10.5 & 4.5 & 4.4 & -\\
      CodeLlama(PAL) & 7B & - & \cmark & \xmark & 27.1 & 17.2 & - & -\\
      DeepSeekMath-Base(PAL) & 7B & - & \cmark & \xmark & 66.9 & 31.4 & - & -\\
      Llama-3           & 8B    & - & \xmark & \xmark & 54.8 & 21.3 & - & -\\
      CodeLlama         & 34B   & - & \xmark & \xmark & 29.6 & 12.2 & 7.0 & -\\
      CodeLlama(PAL)    & 34B & - & \cmark & \xmark & 53.3 & 23.9 & - & -\\
      Llemma            & 34B & - &  \xmark & \xmark & 51.5 & 25.0 & 11.8& -\\
      Llemma(PAL)       & 34B & - & \cmark & \xmark & 62.6 & 27.1 & - & -\\
      \midrule[0.8pt]
      \multicolumn{9}{c}{Single-step reasoning Models} \\
      \midrule[0.8pt]
      WizardMath & Llama2-7B & 96k & \xmark & \cmark & 54.9 & 10.7 & - & -\\
      MAmmoTH-Coder & CodeLlama-7B & 260k & \cmark & \xmark & 59.4 & 33.4 & 11.0 & 15.3\\
      WizardMath & Llama2-70B & 96k & \xmark & \cmark & 81.6& 22.7& - & -\\
      MAmmoTH & Llama2-70B & 260k & \cmark & \xmark & 76.9 & 41.8 & 11.8 & 24.7\\
      \midrule[0.8pt]
      \multicolumn{9}{c}{Multi-step reasoning Models} \\
      \midrule[0.8pt]
      MathCoder & CodeLlama-7B & 80k & \cmark & \cmark & 67.8 & 30.2 & - & -\\
      ToRA-Code & CodeLlama-7B & 69k & \cmark & \cmark & 72.6 & 44.6 & 4.8 & 23.9\\
      MARIO & Llemma-7B & 82k & \cmark & \cmark & 70.1 & 47.0 & 21.7 & 34.5 \\
      MathGenie & Llemma-7B & 250k & \cmark & \cmark & 76.0 & 48.3 & - & - \\
      MathCoder & Llama2-70B & 80k & \cmark & \cmark & 83.9 & 45.1 & - & -\\
      ToRA & Llama2-70B & 69k & \cmark & \cmark & 84.3& 49.7& 9.6 & 30.9\\
      MathGenie & Llama2-70B & 250k & \cmark & \cmark & 88.4 & 51.2 & - & - \\
      \midrule[0.8pt]
      \multicolumn{9}{c}{ \textbf{MCoT reasoning Models}} \\
      \midrule[0.8pt]
      MCoT-DeepSeek & DeepSeekMathBase-7B & 82k$^\dagger$ & \cmark & \cmark & \textbf{78.8} & \textbf{55.8} & \textbf{31.6} & \textbf{41.3} \\
      MCoT-Llemma & Llemma-7B & 82k$^\dagger$ & \cmark & \cmark & 69.3 & 48.1 & 18.0 & 33.3\\
      
      \multirow{2}{*}{MCoT-Llama3} & Llama3-8B & 82k$^\dagger$ & \cmark & \cmark & 76.9 & 47.4 & 8.8 & 32.7 \\
      & Llama3-70B & 82k$^\dagger$ & \cmark & \cmark & 83.1 & 54.7 & 19.9 & 38.7 \\
      \bottomrule[1.2pt]
    \end{tabular}
    }
    \caption{Results on different datasets. The best results of 7B open-source models are bold. $^*$GK2023 represents Gaokao-2023-Math-En dataset. 82k$^\dagger$ represents the count of Markov chains, encompassing approximately 160$k$ step-wise entries. Conversely, data size in prior methods are accounted for by enumerating the complete trajectories of multi-step reasoning.}
    \vspace{-4mm}
    \label{tab:main_exp}
\end{table*}

\paragraph{Accuracy} Table~\ref{tab:main_exp} demonstrates our models outperform other open-source competitive math-solving models, exhibiting a clear advantage across both in-domain datasets and out-of-domain datasets. 
Our model \texttt{MCoT-DeepSeek} is fine-tuned from the DeepSeekMath-Base\textsubscript{7B} on the \texttt{MCoTInstruct} dataset. 
Compared to the base model, our model achieves substantial gains on the GSM8k and MATH datasets, with improvements of about $12\%$ and $24\%$, respectively. 
\texttt{MCoT-DeepSeek} achieve state-of-the-art results across all datasets, in 7B models.

Moreover, our model \texttt{MCoT-DeepSeek} achieves $55.8\%$ on MATH dataset, which surpasses all 34B and 70B models on MATH dataset without any extra strategies, such as majority voting~\citep{wang2022self}. 
Notably, during the training phase in MCoT, the model is exposed only to triplet data like $(\mathbf{q}_{t-1}, \mathbf{s}_{t-1}, \mathbf{q}_{t})$ or $(\mathbf{q}_{T}, \mathbf{s}_{T}, \mathbf{a})$. It has never been trained on the complete solution data. 
However, when a question is presented to the model during the inference stage, it first attempts a single-step solution before deciding whether to further reduce the problem or deliver the final answer.
In the reasoning stage, the model leverages the fragmented knowledge acquired during training to construct a complete Markov chain reasoning process from question to final answer. 


\paragraph{Efficiency} 
To intuitively evaluate the efficiency of different reasoning methods, we compare the performance of multi-step reasoning (MSR) and MCoT reasoning. To ensure a fair comparison, all reasoning approaches utilize an external tool - Python code interpreter. 
The MSR model is fine-tuned on our initial multi-step reasoning dataset, $\mathcal{D}_{\text{origin}}$. 
The MCoT model is fine-tuned on our \texttt{MCoTInstruct} dataset. 
The \texttt{MCoTInstruct} dataset is extended from $\mathcal{D}_{\text{origin}}$, with both the source data and solutions preserved as consistently as possible to minimize the impact of dataset variations. 
Moreover, the maximum number of reasoning steps is set to 8. 

\begin{table*}[htbp]
\setlength{\tabcolsep}{1mm}
\centering
\small
\renewcommand\arraystretch{1.2}

\begin{center}
    \begin{tabular}{llcccc}
        \toprule[1.2pt]  
        \textbf{Base Model} & \textbf{Methods}   & \textbf{GSM8K}($\uparrow$) & \textbf{MATH}($\uparrow$) & \textbf{$E^\S$}($\downarrow$) & \textbf{Cache Memory (GB)$^\dagger$ }($\downarrow$)\\
        \midrule[0.8pt]
            \multirow{2}{*}{DeepSeekMath-Base\textsubscript{7B}} & MSR &  77.3 & 54.9 & 1.12 & 48.7\\
            & MCoT & \textbf{78.8} & \textbf{55.8} & \textbf{0.60} & \textbf{30.2}\\
            \midrule
            \multirow{2}{*}{Llemma\textsubscript{7B}} & MSR & 67.1 & 46.0 & 1.18 & 37.9 \\
            & MCoT & \textbf{69.3}  & \textbf{48.1} & \textbf{0.62} & \textbf{25.7}\\
            \midrule
            \multirow{2}{*}{Llama3\textsubscript{8B}} & MSR & 73.4 & 45.1 & 1.23 & 55.6 \\
            & MCoT & \textbf{76.9} & \textbf{47.4} & \textbf{0.69} & \textbf{37.9}\\
        \bottomrule[1.2pt]
    \end{tabular}
\end{center}
\caption{\label{tab:exp_compare}The comparison between multi-step reasoning (MSR) and MCoT on Accuracy and Efficiency. 
$E^\S$ means the metric defined in Eq.~(\ref{eq:decoding_time}).
$^\dagger$ indicates the average amount of cached GPU memory used per sample.
}
\vspace{-4mm}
\end{table*}

Table~\ref{tab:exp_compare} presents the comparative results of MSR and MCoT regarding reasoning efficiency and accuracy. 
Our observations are as follows: \textbf{(1)} Compared to single-step reasoning in Table~\ref{tab:main_exp}, MSR and MCoT indeed significantly enhance reasoning accuracy. 
\textbf{(2)} Compared to MSR, MCoT demonstrates notable improvements in reasoning accuracy and efficiency. 
For instance, in the Llemma\textsubscript{7B} model, MCoT achieves improvements over MSR on both the MATH (+2.1\%) and GSM8K (+2.2\%) datasets. Furthermore, MCoT's inference efficiency from $E$ is 1.90 times greater than that of MSR. 


\paragraph{Different base models and model sizes}
We evaluated the effectiveness of the MCoT approach using three base models: DeepSeek, Llemma, and the llama3 series. Tables~\ref{tab:main_exp} and~\ref{tab:exp_compare} show that MCoT outperforms multi-step reasoning in accuracy across all models, with significantly smaller cache memory usage.
To investigate the effectiveness across varying model scales, we scale up the Llama3 model size from 8B to 70B, observing notable performance improvements on all benchmarks. 
Specifically, we found that the Llama3 series performs poorly overall on the OCW dataset. 
In our analysis of the OCW dataset, detailed in Appendix~\ref{appendix:test_data}, we found that its format causes Llama3 to mistakenly interpret it as in-context learning, even after fine-tuning with \texttt{MCoTInstruct} dataset.

\subsection{Analysis 1: Efficiency}

\begin{figure*}[htbp]
\centering
\begin{subfigure}[b]{0.4\linewidth}
    \includegraphics[width=\linewidth]{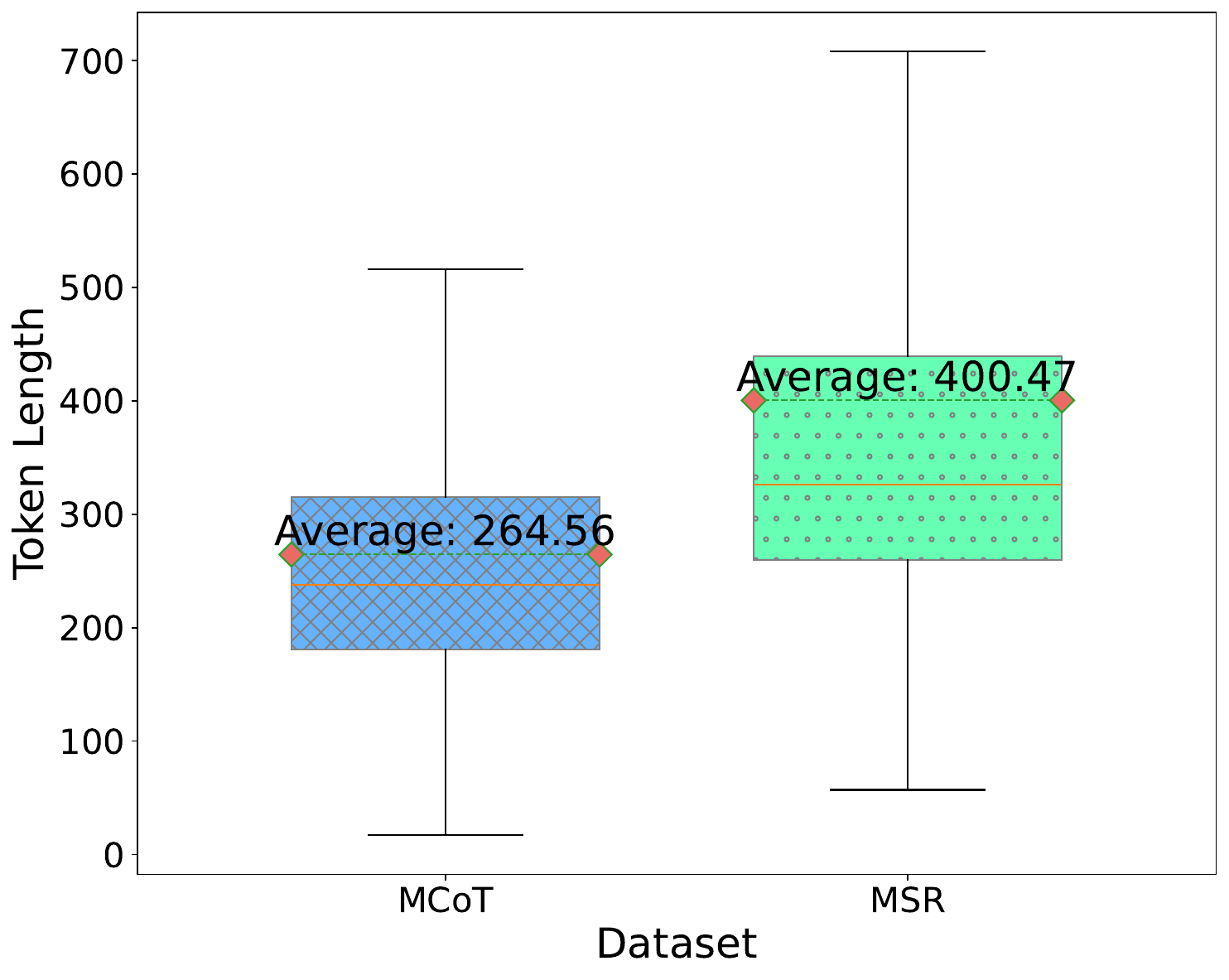}
    \caption{Token length distribution on training }
    \label{fig:exp_token_length}
\end{subfigure}
\hfill
\begin{subfigure}[b]{0.58\linewidth} 
    \includegraphics[width=\linewidth]{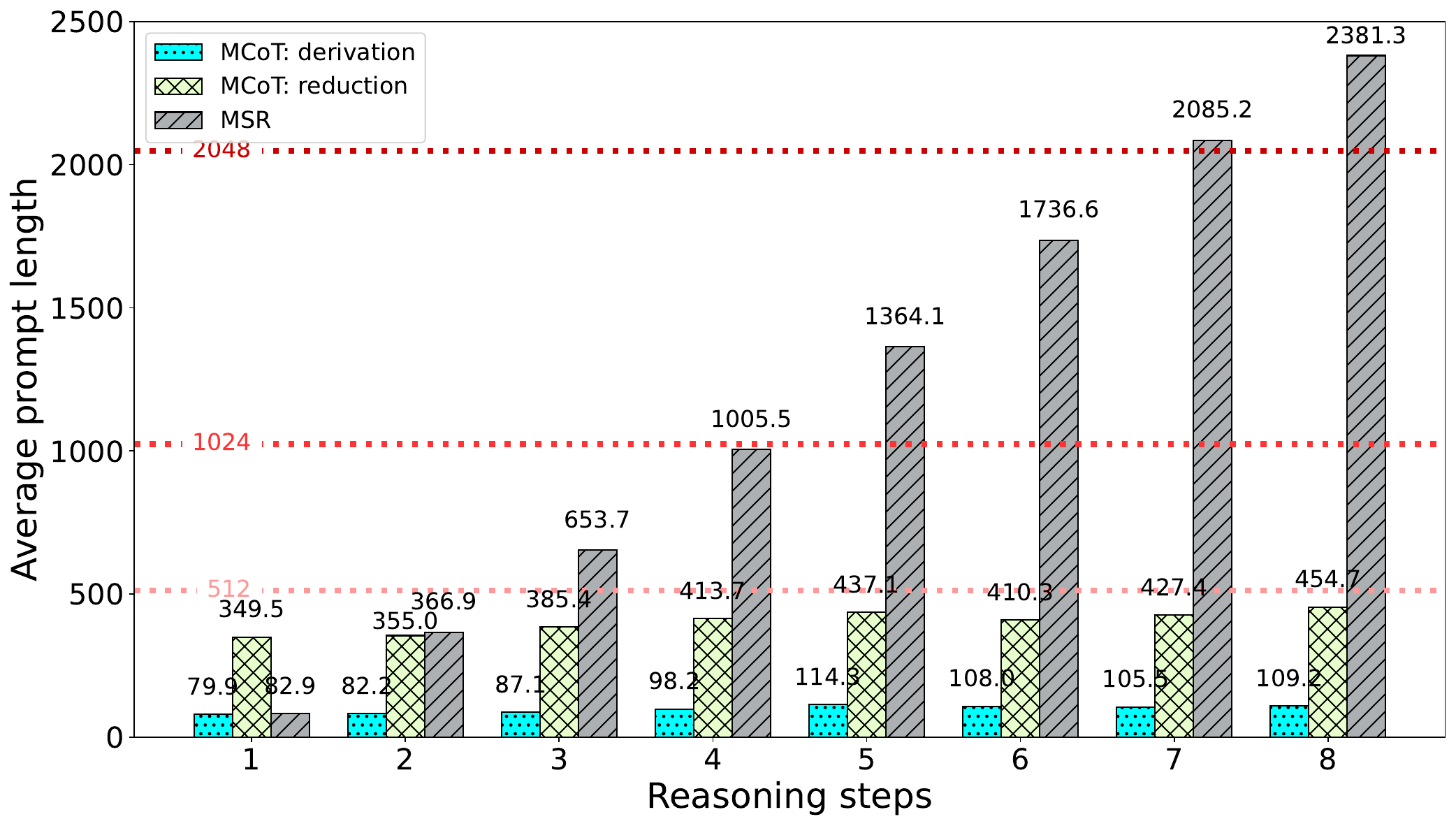} 
    \caption{Average Prompt token Length on test set}
    \label{fig:exp_prompt_lenght}
\end{subfigure}
\vspace{-1mm}
\caption{Comparison of token Length in MCoT and MSR on training and test set.}
\label{fig:exp_prompt_all}
\vspace{-2mm}
\end{figure*}

\paragraph{Efficient Training} Figure~\ref{fig:exp_token_length} illustrates the distribution of token length for the \texttt{MCoTInstruct} dataset and the multi-step reasoning instruction dataset $\mathcal{D}_{\text{origin}}$. In comparison, the token length of MCoT is noticeably shorter than that of MSR, with an average reduction of $135.91$ tokens. The underlying reason is that MCoT is only trained on the $(\mathbf{q}_{t-1}, \mathbf{s}_{t-1}, \mathbf{q}_{t})$ triplets, whereas MSR requires training on the entire trajectory.


\paragraph{Efficient Inference} 
To investigate the efficiency of MCoT during inference, we compare the average prompt length of MCoT and MSR as the number of reasoning steps increases on the MATH test set, using DeepSeekMath-Base as the base model.
As shown in Figure~\ref{fig:exp_prompt_lenght}, there is a stark contrast in the average prompt length between MCoT and MSR. 
MCoT maintains a stable prompt length, unaffected by the increasing reasoning steps, while MSR exhibits a growing prompt length.

In MCoT, the average token length during the derivation stage remains under 128 tokens, while in the reduction stage, it stays below 512 tokens.
This indicates that MCoT can tackle complex mathematical reasoning problems, such as those in the MATH dataset, using only a 512-token context window. 
This significantly reduces memory and computational demands.
In contrast, the average prompt length surpasses 2048 tokens by the seventh step in MSR, indicating a substantial increase in memory and computational requirements as the reasoning process progresses.


\begin{figure*}[t]
\centering
\begin{subfigure}[b]{0.48\linewidth}
    \includegraphics[width=\linewidth]{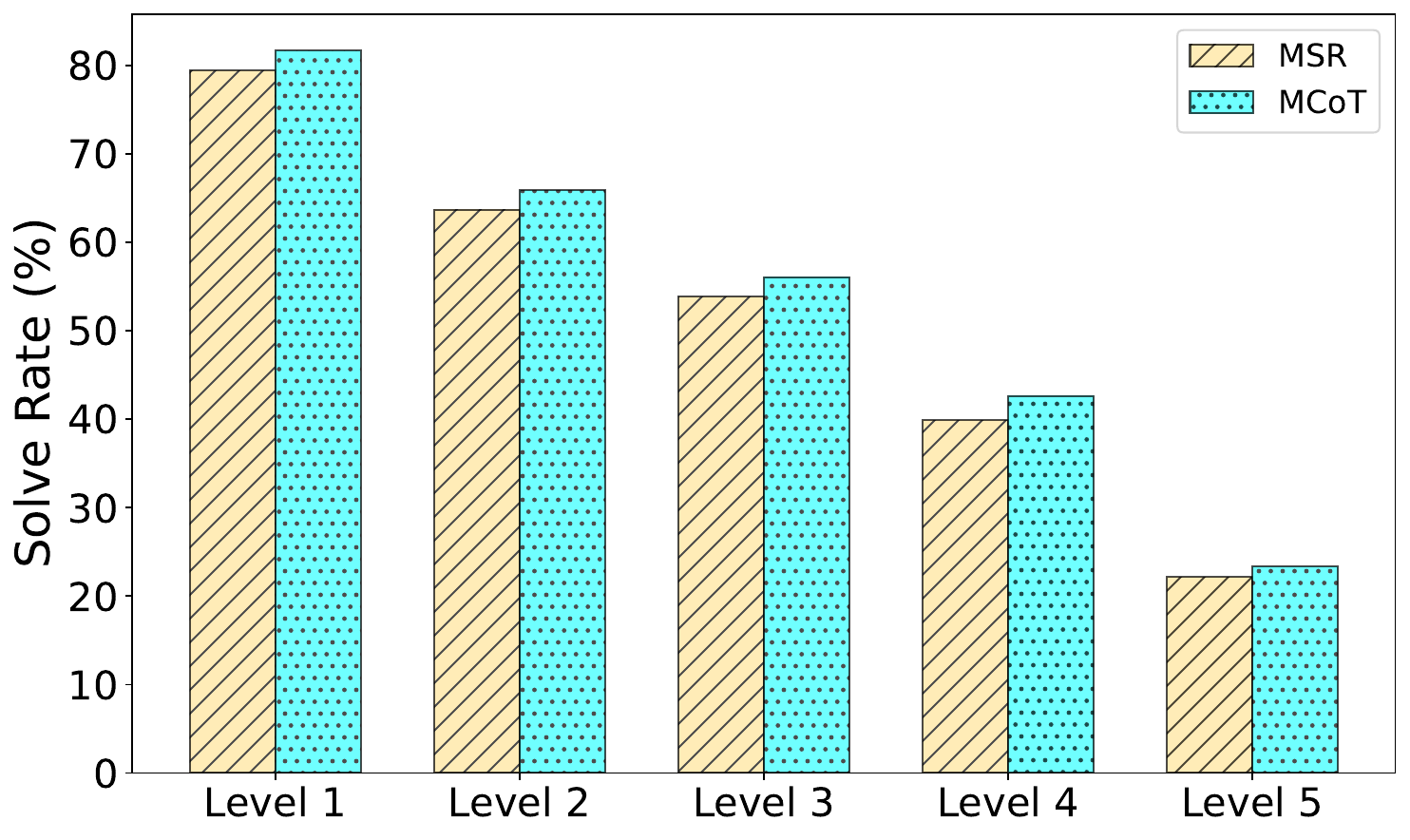}
    \caption{Difficulty Level}
    \label{fig:exp_llemma_math_level}
\end{subfigure}
\hfill
\begin{subfigure}[b]{0.48\linewidth} 
    \includegraphics[width=\linewidth]{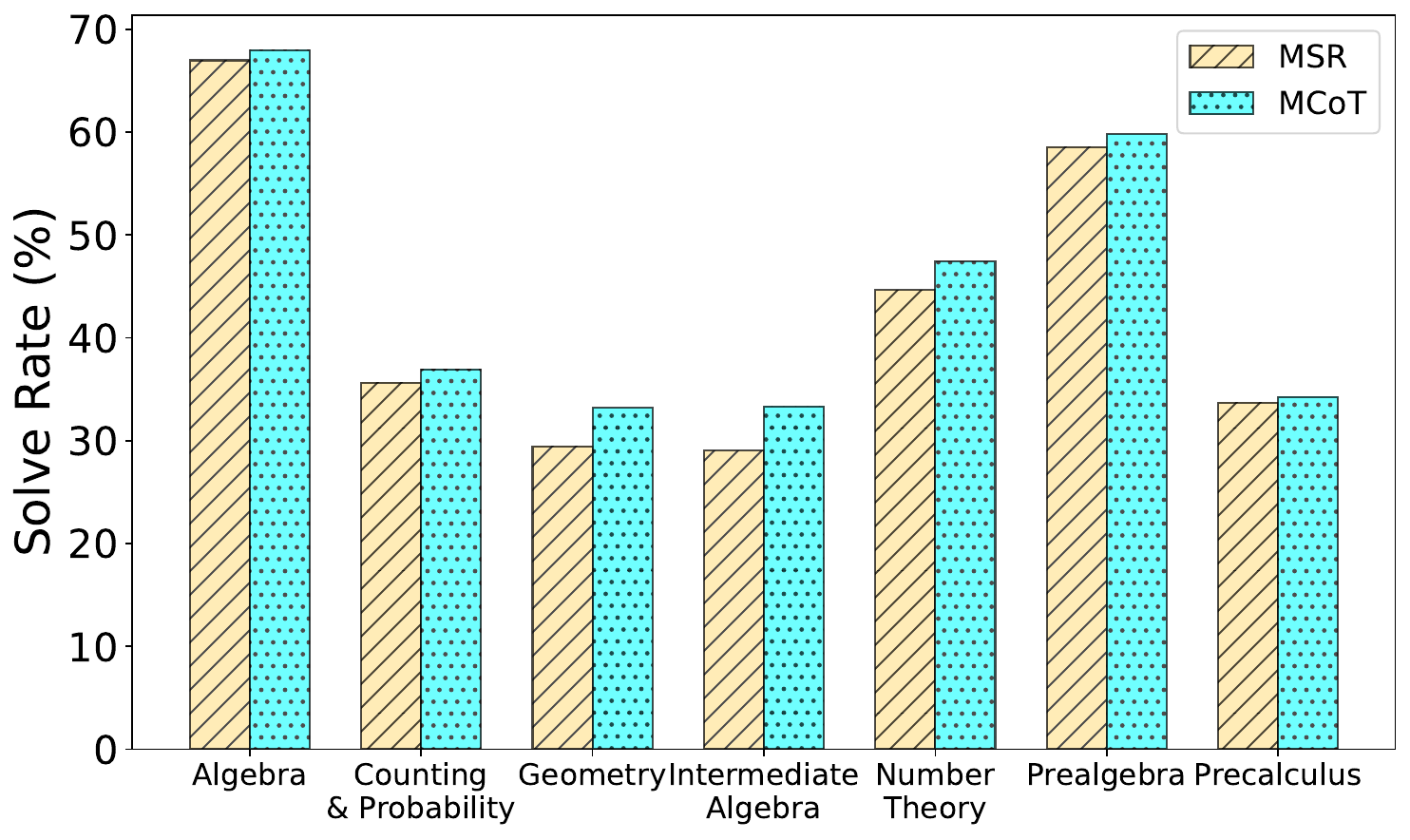} 
    \caption{Different Subject}
    \label{fig:exp_llemma_math_subject}
\end{subfigure}
\vspace{-1mm}
\caption{Comparison of problem solving between MCoT and MSR on MATH test set, with Llemma\textsubscript{7B} as base model.}
\vspace{-4mm}
\label{fig:exp_llemma_math}
\end{figure*}

\subsection{Analysis 2: Problem Solving}

To assess MCoT's problem-solving capabilities, we analyze its performance on the MATH test set across various difficulty levels and subjects, calculating the success rate for each category.
In Figure~\ref{fig:exp_llemma_math_level}, it is demonstrated that MCoT achieves a higher success rate in solving more challenging problems compared to MSR. 
We attribute this superior performance to MCoT's training method, which emphasizes derivation and reduction techniques rather than the complete reasoning path from question to answer. This approach enhances the model's generalization ability through autonomous and iterative problem-solving.
Figure~\ref{fig:exp_llemma_math_subject} shows that MCoT consistently excels in solving a broader range of problems across various subjects. 
More details and results \emph{w.r.t.} other base models, such as Deepseek and Llama, are shown in Figure~\ref{fig:exp_deepseek_math} and \ref{fig:exp_llama3_math} in Appendix~\ref{appendix:analysis}.

\subsection{Analysis 3: Hybrid training strategy}


The MCoT approach notes that the model is not trained on full solution datasets, limiting its ability to generate complete solutions independently. 
To address this, we explore a hybrid training method that partially exposes the model to full solutions to assess performance improvements. 
We used the DeepSeekMath-Base-7B model, starting with a 2-epoch warm-up on a 27.4k multi-step reasoning dataset from GSM8K and MATH, followed by training on the MCoTInstruct dataset.

Table~\ref{tab:hybrid_training} shows that the hybrid training improves in-domain performance but performs worse on OOD datasets compared to direct MCoT training. 
This may be due to overfitting caused by the inclusion of complete solution data, leading to weaker generalization. 
We will leave more exploration on how to train the LLM with mixed data as future work.

\begin{table}[!ht]
\footnotesize
\setlength{\tabcolsep}{2pt}
\renewcommand\arraystretch{1.2}
\centering
\begin{tabular}{lcccc}
\toprule[1.2pt]
       & GSM8K & MATH & OCW & GK2023 \\
\midrule[0.8pt]
DeepSeekMath-Base & 66.90 & 31.40 & - & - \\
\hline
MSR-DeepSeek & 77.30 & 54.90 & 27.94  & 38.96  \\
MCoT-DeepSeek & 78.80 & 55.80 & \textbf{31.60} & \textbf{41.30} \\
\rowcolor{lightgray}
Hybrid Training & & & &\\
MSR-DeepSeek\textsubscript{warm-up} & 75.97 & 52.12 & 25.37 & 37.40 \\
~~+ Continue \emph{w.r.t.} MCoT & \textbf{79.30} & \textbf{56.60} & 29.41 & 39.48 \\
\bottomrule[1.2pt]
\end{tabular}
\caption{The results on hybrid training strategy within DeepSeekMath-Base\textsubscript{7B} model.}
\label{tab:hybrid_training}
\vspace{-4mm}
\end{table}

\subsection{Case study: Self-correction in MCoT}
We provide a detailed case study in Appendix~\ref{appendix:case} to illustrate the reasoning process, highlighting MCoT's self-correction capabilities. Our experiments demonstrate that MCoT does not empirically propagate errors more significantly than traditional MSR frameworks. This is largely because MCoT incorporates a self-correction mechanism, as illustrated in Figure~\ref{fig:self_correction} and case study in Appendix~\ref{appendix:case}. Specifically, MCoT can summarize prior errors into the context of the next reasoning step and address them without needing to retain the full historical context.







\section{Related Work}
\textbf{Chain of Reasoning}
LLMs have exhibited strong reasoning capabilities by utilizing Chain of Thought~\citep{wei2022chain, brown2020language} prompting. 
Tree of Thoughts (ToT)~\citep{yao2024tree} enables exploration over coherent units of thoughts that serve as intermediate steps toward problem-solving. Program of Thought (PoT)~\citep{chen2022program} enhances the capabilities of LLMs to use programs as thought processes. 
Several works~\citep{zhou2022least, wang2022iteratively, li2023making, wang2023making, wang2023plan} have developed CoT or PoT technology to employ LLMs to tackle reasoning tasks by allowing intermediate steps. 
It is important to note that in these methods, the intermediate steps are preserved as historical context, making them dependent. In contrast, our approach leverages the inherent independence of the Markov chain to separate the intermediate steps.

\textbf{Mathematical Reasoning}
Recent works~\citep{wang2023mathcoder, gou2023tora, liao2024mario, lu2024mathgenie} have made significant advancements in enhancing reasoning capabilities within LLMs through the implementation of step-by-step natural language reasoning, achieving better results than single-step reasoning~\citep{luo2023wizardmath, yu2023metamath, yue2023mammoth}. 
In single-step reasoning, \cite{luo2023wizardmath} and \cite{yu2023metamath} utilize textual content as solutions, and \cite{yue2023mammoth} introduces a unique hybrid of CoT and PoT rationales. 
In multi-step reasoning, \citet{wang2023mathcoder} and \citet{gou2023tora} incorporate code snippets or tools within each step of the reasoning process, \citet{liao2024mario} also adds text analysis based on the code snippets at each step. 
Nevertheless, as the number of reasoning steps increases, the multi-step reasoning involves processing long contexts demands greater computational resources and a decline in reasoning drops~\citep{levy2024same}.
In our work, we utilize a process of \textit{derivation, then reduction} in MCoT. This reduction step simplifies the historical questions and solutions into independent questions, enhancing clarity and efficiency.

\section{Conclusion}
This paper presents MCoT, an innovative Markov Chain of Thought framework for efficient multi-step reasoning. 
Our framework leverages the independence of Markov chains, conceptualizing the solution process as a series of state transitions. This approach enables LLMs to address complex reasoning tasks more efficiently and intelligently.
MCoT achieves superior performance on diverse mathematical reasoning tasks, substantially outperforming existing multi-step reasoning approaches in both efficiency and accuracy. 
Our work provides a new pathway for solving complex reasoning tasks.

\section*{Limitations}
The limitation of Markov Chain of Thought (MCoT) models primarily arises when an error occurs at an intermediate step, which can cascade and lead to failures in subsequent steps. This is due to the Markov property, which assumes that each step depends only on the current state, rather than on the sequence of events that preceded it. Consequently, errors are not corrected and may propagate through the chain, resulting in flawed conclusions.

To address this limitation, integrating Monte Carlo Tree Search (MCTS)~\citep{browne2012survey, silver2016mastering, silver2017mastering} could be a robust solution. By integrating MCTS with MCoT, we equip the reasoning models with the capacity for backtracking and learning from simulated explorations. 
This synergy not only addresses the original limitation of the MCoT concerning the independence assumption but also fortifies the model's ability to navigate complex problems more effectively. This potential direction is what we intend to explore in future work.

\section*{Ethical Considerations}
The MCoT framework introduces a novel paradigm for efficiently and intelligently handling complex reasoning tasks. 
Currently, our focus is on mathematical reasoning, Therefore, this work does not have direct negative social impacts.
In our experiments, we used publicly available datasets widely employed in prior research, containing no sensitive information to the best of our knowledge. The authors have followed ACL ethical guidelines, and the application of this work poses no apparent ethical risks.

\section*{Acknowledgement}

This work was supported by Alibaba Research Intern Program.

\bibliography{reference}

\begin{thebibliography}{56}
\providecommand{\natexlab}[1]{#1}

\bibitem[{Achiam et~al.(2023)Achiam, Adler, Agarwal, Ahmad, Akkaya, Aleman, Almeida, Altenschmidt, Altman, Anadkat et~al.}]{achiam2023gpt}
Josh Achiam, Steven Adler, Sandhini Agarwal, Lama Ahmad, Ilge Akkaya, Florencia~Leoni Aleman, Diogo Almeida, Janko Altenschmidt, Sam Altman, Shyamal Anadkat, et~al. 2023.
\newblock Gpt-4 technical report.
\newblock \emph{arXiv preprint arXiv:2303.08774}.

\bibitem[{AI@Meta(2024)}]{llamma3blog}
AI@Meta. 2024.
\newblock \href {https://ai.meta.com/blog/meta-llama-3/} {{I}ntroducing {M}eta {L}lama 3: The most capable openly available {LLM} to date}.

\bibitem[{Anil et~al.(2023)Anil, Dai, Firat, Johnson, Lepikhin, Passos, Shakeri, Taropa, Bailey, Chen et~al.}]{anil2023palm}
Rohan Anil, Andrew~M Dai, Orhan Firat, Melvin Johnson, Dmitry Lepikhin, Alexandre Passos, Siamak Shakeri, Emanuel Taropa, Paige Bailey, Zhifeng Chen, et~al. 2023.
\newblock Palm 2 technical report.
\newblock \emph{arXiv preprint arXiv:2305.10403}.

\bibitem[{Azerbayev et~al.(2023)Azerbayev, Schoelkopf, Paster, Santos, McAleer, Jiang, Deng, Biderman, and Welleck}]{azerbayev2023llemma}
Zhangir Azerbayev, Hailey Schoelkopf, Keiran Paster, Marco~Dos Santos, Stephen McAleer, Albert~Q Jiang, Jia Deng, Stella Biderman, and Sean Welleck. 2023.
\newblock Llemma: An open language model for mathematics.
\newblock \emph{arXiv preprint arXiv:2310.10631}.

\bibitem[{Brown et~al.(2020)Brown, Mann, Ryder, Subbiah, Kaplan, Dhariwal, Neelakantan, Shyam, Sastry, Askell et~al.}]{brown2020language}
Tom Brown, Benjamin Mann, Nick Ryder, Melanie Subbiah, Jared~D Kaplan, Prafulla Dhariwal, Arvind Neelakantan, Pranav Shyam, Girish Sastry, Amanda Askell, et~al. 2020.
\newblock Language models are few-shot learners.
\newblock \emph{Advances in neural information processing systems}, 33:1877--1901.

\bibitem[{Browne et~al.(2012)Browne, Powley, Whitehouse, Lucas, Cowling, Rohlfshagen, Tavener, Perez, Samothrakis, and Colton}]{browne2012survey}
Cameron~B Browne, Edward Powley, Daniel Whitehouse, Simon~M Lucas, Peter~I Cowling, Philipp Rohlfshagen, Stephen Tavener, Diego Perez, Spyridon Samothrakis, and Simon Colton. 2012.
\newblock A survey of monte carlo tree search methods.
\newblock \emph{IEEE Transactions on Computational Intelligence and AI in games}, 4(1):1--43.

\bibitem[{Caciularu et~al.(2023)Caciularu, Peters, Goldberger, Dagan, and Cohan}]{caciularu2023peek}
Avi Caciularu, Matthew~E Peters, Jacob Goldberger, Ido Dagan, and Arman Cohan. 2023.
\newblock Peek across: Improving multi-document modeling via cross-document question-answering.
\newblock \emph{arXiv preprint arXiv:2305.15387}.

\bibitem[{Chen et~al.(2024{\natexlab{a}})Chen, Liao, Li, and Fan}]{chen2024alphamath}
Guoxin Chen, Minpeng Liao, Chengxi Li, and Kai Fan. 2024{\natexlab{a}}.
\newblock Alphamath almost zero: process supervision without process.
\newblock \emph{arXiv preprint arXiv:2405.03553}.

\bibitem[{Chen et~al.(2024{\natexlab{b}})Chen, Liao, Li, and Fan}]{chen2024step}
Guoxin Chen, Minpeng Liao, Chengxi Li, and Kai Fan. 2024{\natexlab{b}}.
\newblock Step-level value preference optimization for mathematical reasoning.
\newblock \emph{arXiv preprint arXiv:2406.10858}.

\bibitem[{Chen et~al.(2022)Chen, Ma, Wang, and Cohen}]{chen2022program}
Wenhu Chen, Xueguang Ma, Xinyi Wang, and William~W Cohen. 2022.
\newblock Program of thoughts prompting: Disentangling computation from reasoning for numerical reasoning tasks.
\newblock \emph{arXiv preprint arXiv:2211.12588}.

\bibitem[{Chen et~al.(2023)Chen, Qian, Tang, Lai, Liu, Han, and Jia}]{chen2023longlora}
Yukang Chen, Shengju Qian, Haotian Tang, Xin Lai, Zhijian Liu, Song Han, and Jiaya Jia. 2023.
\newblock Longlora: Efficient fine-tuning of long-context large language models.
\newblock \emph{arXiv preprint arXiv:2309.12307}.

\bibitem[{Cobbe et~al.(2021)Cobbe, Kosaraju, Bavarian, Chen, Jun, Kaiser, Plappert, Tworek, Hilton, Nakano et~al.}]{cobbe2021training}
Karl Cobbe, Vineet Kosaraju, Mohammad Bavarian, Mark Chen, Heewoo Jun, Lukasz Kaiser, Matthias Plappert, Jerry Tworek, Jacob Hilton, Reiichiro Nakano, et~al. 2021.
\newblock Training verifiers to solve math word problems.
\newblock \emph{arXiv preprint arXiv:2110.14168}.

\bibitem[{Dao(2023)}]{dao2023flashattention}
Tri Dao. 2023.
\newblock Flashattention-2: Faster attention with better parallelism and work partitioning.
\newblock \emph{arXiv preprint arXiv:2307.08691}.

\bibitem[{Dua et~al.(2022)Dua, Gupta, Singh, and Gardner}]{dua2022successive}
Dheeru Dua, Shivanshu Gupta, Sameer Singh, and Matt Gardner. 2022.
\newblock Successive prompting for decomposing complex questions.
\newblock \emph{arXiv preprint arXiv:2212.04092}.

\bibitem[{Gao et~al.(2023{\natexlab{a}})Gao, Madaan, Zhou, Alon, Liu, Yang, Callan, and Neubig}]{gao2023pal}
Luyu Gao, Aman Madaan, Shuyan Zhou, Uri Alon, Pengfei Liu, Yiming Yang, Jamie Callan, and Graham Neubig. 2023{\natexlab{a}}.
\newblock Pal: Program-aided language models.
\newblock In \emph{International Conference on Machine Learning}, pages 10764--10799. PMLR.

\bibitem[{Gao et~al.(2023{\natexlab{b}})Gao, Xiong, Gao, Jia, Pan, Bi, Dai, Sun, and Wang}]{gao2023retrieval}
Yunfan Gao, Yun Xiong, Xinyu Gao, Kangxiang Jia, Jinliu Pan, Yuxi Bi, Yi~Dai, Jiawei Sun, and Haofen Wang. 2023{\natexlab{b}}.
\newblock Retrieval-augmented generation for large language models: A survey.
\newblock \emph{arXiv preprint arXiv:2312.10997}.

\bibitem[{Gou et~al.(2023)Gou, Shao, Gong, Yang, Huang, Duan, Chen et~al.}]{gou2023tora}
Zhibin Gou, Zhihong Shao, Yeyun Gong, Yujiu Yang, Minlie Huang, Nan Duan, Weizhu Chen, et~al. 2023.
\newblock Tora: A tool-integrated reasoning agent for mathematical problem solving.
\newblock \emph{arXiv preprint arXiv:2309.17452}.

\bibitem[{Gugger et~al.(2022)Gugger, Debut, Wolf, Schmid, Mueller, Mangrulkar, Sun, and Bossan}]{accelerate}
Sylvain Gugger, Lysandre Debut, Thomas Wolf, Philipp Schmid, Zachary Mueller, Sourab Mangrulkar, Marc Sun, and Benjamin Bossan. 2022.
\newblock Accelerate: Training and inference at scale made simple, efficient and adaptable.
\newblock \url{https://github.com/huggingface/accelerate}.

\bibitem[{Hendrycks et~al.(2021)Hendrycks, Burns, Kadavath, Arora, Basart, Tang, Song, and Steinhardt}]{hendrycks2021measuring}
Dan Hendrycks, Collin Burns, Saurav Kadavath, Akul Arora, Steven Basart, Eric Tang, Dawn Song, and Jacob Steinhardt. 2021.
\newblock Measuring mathematical problem solving with the math dataset.
\newblock \emph{arXiv preprint arXiv:2103.03874}.

\bibitem[{Huang et~al.(2023)Huang, Cheng, Shu, Bao, and Qu}]{huang2023question}
Xiang Huang, Sitao Cheng, Yiheng Shu, Yuheng Bao, and Yuzhong Qu. 2023.
\newblock Question decomposition tree for answering complex questions over knowledge bases.
\newblock In \emph{Proceedings of the AAAI Conference on Artificial Intelligence}, volume~37, pages 12924--12932.

\bibitem[{Kingma and Ba(2014)}]{kingma2014adam}
Diederik~P Kingma and Jimmy Ba. 2014.
\newblock Adam: A method for stochastic optimization.
\newblock \emph{arXiv preprint arXiv:1412.6980}.

\bibitem[{Levy et~al.(2024)Levy, Jacoby, and Goldberg}]{levy2024same}
Mosh Levy, Alon Jacoby, and Yoav Goldberg. 2024.
\newblock Same task, more tokens: the impact of input length on the reasoning performance of large language models.
\newblock \emph{arXiv preprint arXiv:2402.14848}.

\bibitem[{Lewis et~al.(2020)Lewis, Perez, Piktus, Petroni, Karpukhin, Goyal, K{\"u}ttler, Lewis, Yih, Rockt{\"a}schel et~al.}]{lewis2020retrieval}
Patrick Lewis, Ethan Perez, Aleksandra Piktus, Fabio Petroni, Vladimir Karpukhin, Naman Goyal, Heinrich K{\"u}ttler, Mike Lewis, Wen-tau Yih, Tim Rockt{\"a}schel, et~al. 2020.
\newblock Retrieval-augmented generation for knowledge-intensive nlp tasks.
\newblock \emph{Advances in Neural Information Processing Systems}, 33:9459--9474.

\bibitem[{Lewkowycz et~al.(2022)Lewkowycz, Andreassen, Dohan, Dyer, Michalewski, Ramasesh, Slone, Anil, Schlag, Gutman-Solo et~al.}]{lewkowycz2022solving}
Aitor Lewkowycz, Anders Andreassen, David Dohan, Ethan Dyer, Henryk Michalewski, Vinay Ramasesh, Ambrose Slone, Cem Anil, Imanol Schlag, Theo Gutman-Solo, et~al. 2022.
\newblock Solving quantitative reasoning problems with language models.
\newblock \emph{Advances in Neural Information Processing Systems}, 35:3843--3857.

\bibitem[{Li et~al.(2023)Li, Lin, Zhang, Fu, Chen, Lou, and Chen}]{li2023making}
Yifei Li, Zeqi Lin, Shizhuo Zhang, Qiang Fu, Bei Chen, Jian-Guang Lou, and Weizhu Chen. 2023.
\newblock Making language models better reasoners with step-aware verifier.
\newblock In \emph{Proceedings of the 61st Annual Meeting of the Association for Computational Linguistics (Volume 1: Long Papers)}, pages 5315--5333.

\bibitem[{Liao et~al.(2024)Liao, Luo, Li, Wu, and Fan}]{liao2024mario}
Minpeng Liao, Wei Luo, Chengxi Li, Jing Wu, and Kai Fan. 2024.
\newblock Mario: Math reasoning with code interpreter output--a reproducible pipeline.
\newblock \emph{arXiv preprint arXiv:2401.08190}.

\bibitem[{Lu et~al.(2024)Lu, Zhou, Ren, Wang, Shi, Pan, Zhan, and Li}]{lu2024mathgenie}
Zimu Lu, Aojun Zhou, Houxing Ren, Ke~Wang, Weikang Shi, Junting Pan, Mingjie Zhan, and Hongsheng Li. 2024.
\newblock Mathgenie: Generating synthetic data with question back-translation for enhancing mathematical reasoning of llms.
\newblock \emph{arXiv preprint arXiv:2402.16352}.

\bibitem[{Luo et~al.(2023)Luo, Sun, Xu, Zhao, Lou, Tao, Geng, Lin, Chen, and Zhang}]{luo2023wizardmath}
Haipeng Luo, Qingfeng Sun, Can Xu, Pu~Zhao, Jianguang Lou, Chongyang Tao, Xiubo Geng, Qingwei Lin, Shifeng Chen, and Dongmei Zhang. 2023.
\newblock Wizardmath: Empowering mathematical reasoning for large language models via reinforced evol-instruct.
\newblock \emph{arXiv preprint arXiv:2308.09583}.

\bibitem[{Meadows(2008)}]{meadows2008thinking}
Donella~H Meadows. 2008.
\newblock \emph{Thinking in systems: A primer}.
\newblock chelsea green publishing.

\bibitem[{Moura and Ullrich(2021)}]{moura2021lean}
Leonardo~de Moura and Sebastian Ullrich. 2021.
\newblock The lean 4 theorem prover and programming language.
\newblock In \emph{Automated Deduction--CADE 28: 28th International Conference on Automated Deduction, Virtual Event, July 12--15, 2021, Proceedings 28}, pages 625--635. Springer.

\bibitem[{Ouyang et~al.(2022)Ouyang, Wu, Jiang, Almeida, Wainwright, Mishkin, Zhang, Agarwal, Slama, Ray et~al.}]{ouyang2022training}
Long Ouyang, Jeffrey Wu, Xu~Jiang, Diogo Almeida, Carroll Wainwright, Pamela Mishkin, Chong Zhang, Sandhini Agarwal, Katarina Slama, Alex Ray, et~al. 2022.
\newblock Training language models to follow instructions with human feedback.
\newblock \emph{Advances in neural information processing systems}, 35:27730--27744.

\bibitem[{Polya(2004)}]{polya2004solve}
George Polya. 2004.
\newblock \emph{How to solve it: A new aspect of mathematical method}.
\newblock 246. Princeton university press.

\bibitem[{Radhakrishnan et~al.(2023)Radhakrishnan, Nguyen, Chen, Chen, Denison, Hernandez, Durmus, Hubinger, Kernion, Luko{\v{s}}i{\=u}t{\.e} et~al.}]{radhakrishnan2023question}
Ansh Radhakrishnan, Karina Nguyen, Anna Chen, Carol Chen, Carson Denison, Danny Hernandez, Esin Durmus, Evan Hubinger, Jackson Kernion, Kamil{\.e} Luko{\v{s}}i{\=u}t{\.e}, et~al. 2023.
\newblock Question decomposition improves the faithfulness of model-generated reasoning.
\newblock \emph{arXiv preprint arXiv:2307.11768}.

\bibitem[{Rajbhandari et~al.(2021)Rajbhandari, Ruwase, Rasley, Smith, and He}]{rajbhandari2021zero}
Samyam Rajbhandari, Olatunji Ruwase, Jeff Rasley, Shaden Smith, and Yuxiong He. 2021.
\newblock Zero-infinity: Breaking the gpu memory wall for extreme scale deep learning.
\newblock In \emph{Proceedings of the international conference for high performance computing, networking, storage and analysis}, pages 1--14.

\bibitem[{Roziere et~al.(2023)Roziere, Gehring, Gloeckle, Sootla, Gat, Tan, Adi, Liu, Remez, Rapin et~al.}]{roziere2023code}
Baptiste Roziere, Jonas Gehring, Fabian Gloeckle, Sten Sootla, Itai Gat, Xiaoqing~Ellen Tan, Yossi Adi, Jingyu Liu, Tal Remez, J{\'e}r{\'e}my Rapin, et~al. 2023.
\newblock Code llama: Open foundation models for code.
\newblock \emph{arXiv preprint arXiv:2308.12950}.

\bibitem[{Shao et~al.(2024)Shao, Wang, Zhu, Xu, Song, Zhang, Li, Wu, and Guo}]{shao2024deepseekmath}
Zhihong Shao, Peiyi Wang, Qihao Zhu, Runxin Xu, Junxiao Song, Mingchuan Zhang, YK~Li, Y~Wu, and Daya Guo. 2024.
\newblock Deepseekmath: Pushing the limits of mathematical reasoning in open language models.
\newblock \emph{arXiv preprint arXiv:2402.03300}.

\bibitem[{Silver et~al.(2016)Silver, Huang, Maddison, Guez, Sifre, Van Den~Driessche, Schrittwieser, Antonoglou, Panneershelvam, Lanctot et~al.}]{silver2016mastering}
David Silver, Aja Huang, Chris~J Maddison, Arthur Guez, Laurent Sifre, George Van Den~Driessche, Julian Schrittwieser, Ioannis Antonoglou, Veda Panneershelvam, Marc Lanctot, et~al. 2016.
\newblock Mastering the game of go with deep neural networks and tree search.
\newblock \emph{nature}, 529(7587):484--489.

\bibitem[{Silver et~al.(2017)Silver, Schrittwieser, Simonyan, Antonoglou, Huang, Guez, Hubert, Baker, Lai, Bolton et~al.}]{silver2017mastering}
David Silver, Julian Schrittwieser, Karen Simonyan, Ioannis Antonoglou, Aja Huang, Arthur Guez, Thomas Hubert, Lucas Baker, Matthew Lai, Adrian Bolton, et~al. 2017.
\newblock Mastering the game of go without human knowledge.
\newblock \emph{nature}, 550(7676):354--359.

\bibitem[{Simon and Newell(1971)}]{simon1971human}
Herbert~A Simon and Allen Newell. 1971.
\newblock Human problem solving: The state of the theory in 1970.
\newblock \emph{American psychologist}, 26(2):145.

\bibitem[{Taori et~al.(2023)Taori, Gulrajani, Zhang, Dubois, Li, Guestrin, Liang, and Hashimoto}]{taori2023stanford}
Rohan Taori, Ishaan Gulrajani, Tianyi Zhang, Yann Dubois, Xuechen Li, Carlos Guestrin, Percy Liang, and Tatsunori~B Hashimoto. 2023.
\newblock Stanford alpaca: An instruction-following llama model.

\bibitem[{Touvron et~al.(2023)Touvron, Martin, Stone, Albert, Almahairi, Babaei, Bashlykov, Batra, Bhargava, Bhosale et~al.}]{touvron2023llama2}
Hugo Touvron, Louis Martin, Kevin Stone, Peter Albert, Amjad Almahairi, Yasmine Babaei, Nikolay Bashlykov, Soumya Batra, Prajjwal Bhargava, Shruti Bhosale, et~al. 2023.
\newblock Llama 2: Open foundation and fine-tuned chat models.
\newblock \emph{arXiv preprint arXiv:2307.09288}.

\bibitem[{Wang et~al.(2022{\natexlab{a}})Wang, Deng, and Sun}]{wang2022iteratively}
Boshi Wang, Xiang Deng, and Huan Sun. 2022{\natexlab{a}}.
\newblock Iteratively prompt pre-trained language models for chain of thought.
\newblock \emph{arXiv preprint arXiv:2203.08383}.

\bibitem[{Wang et~al.(2023{\natexlab{a}})Wang, Ren, Zhou, Lu, Luo, Shi, Zhang, Song, Zhan, and Li}]{wang2023mathcoder}
Ke~Wang, Houxing Ren, Aojun Zhou, Zimu Lu, Sichun Luo, Weikang Shi, Renrui Zhang, Linqi Song, Mingjie Zhan, and Hongsheng Li. 2023{\natexlab{a}}.
\newblock Mathcoder: Seamless code integration in llms for enhanced mathematical reasoning.
\newblock \emph{arXiv preprint arXiv:2310.03731}.

\bibitem[{Wang et~al.(2023{\natexlab{b}})Wang, Xu, Lan, Hu, Lan, Lee, and Lim}]{wang2023plan}
Lei Wang, Wanyu Xu, Yihuai Lan, Zhiqiang Hu, Yunshi Lan, Roy Ka-Wei Lee, and Ee-Peng Lim. 2023{\natexlab{b}}.
\newblock Plan-and-solve prompting: Improving zero-shot chain-of-thought reasoning by large language models.
\newblock \emph{arXiv preprint arXiv:2305.04091}.

\bibitem[{Wang et~al.(2023{\natexlab{c}})Wang, Li, Chen, Song, Lin, Cao, Liu, and Sui}]{wang2023making}
Peiyi Wang, Lei Li, Liang Chen, Feifan Song, Binghuai Lin, Yunbo Cao, Tianyu Liu, and Zhifang Sui. 2023{\natexlab{c}}.
\newblock Making large language models better reasoners with alignment.
\newblock \emph{arXiv preprint arXiv:2309.02144}.

\bibitem[{Wang et~al.(2022{\natexlab{b}})Wang, Wei, Schuurmans, Le, Chi, Narang, Chowdhery, and Zhou}]{wang2022self}
Xuezhi Wang, Jason Wei, Dale Schuurmans, Quoc Le, Ed~Chi, Sharan Narang, Aakanksha Chowdhery, and Denny Zhou. 2022{\natexlab{b}}.
\newblock Self-consistency improves chain of thought reasoning in language models.
\newblock \emph{arXiv preprint arXiv:2203.11171}.

\bibitem[{Wei et~al.(2022)Wei, Wang, Schuurmans, Bosma, Xia, Chi, Le, Zhou et~al.}]{wei2022chain}
Jason Wei, Xuezhi Wang, Dale Schuurmans, Maarten Bosma, Fei Xia, Ed~Chi, Quoc~V Le, Denny Zhou, et~al. 2022.
\newblock Chain-of-thought prompting elicits reasoning in large language models.
\newblock \emph{Advances in neural information processing systems}, 35:24824--24837.

\bibitem[{Xiong et~al.(2023)Xiong, Liu, Molybog, Zhang, Bhargava, Hou, Martin, Rungta, Sankararaman, Oguz et~al.}]{xiong2023effective}
Wenhan Xiong, Jingyu Liu, Igor Molybog, Hejia Zhang, Prajjwal Bhargava, Rui Hou, Louis Martin, Rashi Rungta, Karthik~Abinav Sankararaman, Barlas Oguz, et~al. 2023.
\newblock Effective long-context scaling of foundation models.
\newblock \emph{arXiv preprint arXiv:2309.16039}.

\bibitem[{Yang and Klein(2021)}]{yang2021fudge}
Kevin Yang and Dan Klein. 2021.
\newblock Fudge: Controlled text generation with future discriminators.
\newblock \emph{arXiv preprint arXiv:2104.05218}.

\bibitem[{Yao et~al.(2024)Yao, Yu, Zhao, Shafran, Griffiths, Cao, and Narasimhan}]{yao2024tree}
Shunyu Yao, Dian Yu, Jeffrey Zhao, Izhak Shafran, Tom Griffiths, Yuan Cao, and Karthik Narasimhan. 2024.
\newblock Tree of thoughts: Deliberate problem solving with large language models.
\newblock \emph{Advances in Neural Information Processing Systems}, 36.

\bibitem[{Yao et~al.(2022)Yao, Zhao, Yu, Du, Shafran, Narasimhan, and Cao}]{yao2022react}
Shunyu Yao, Jeffrey Zhao, Dian Yu, Nan Du, Izhak Shafran, Karthik~R Narasimhan, and Yuan Cao. 2022.
\newblock React: Synergizing reasoning and acting in language models.
\newblock In \emph{The Eleventh International Conference on Learning Representations}.

\bibitem[{Yu et~al.(2023)Yu, Jiang, Shi, Yu, Liu, Zhang, Kwok, Li, Weller, and Liu}]{yu2023metamath}
Longhui Yu, Weisen Jiang, Han Shi, Jincheng Yu, Zhengying Liu, Yu~Zhang, James~T Kwok, Zhenguo Li, Adrian Weller, and Weiyang Liu. 2023.
\newblock Metamath: Bootstrap your own mathematical questions for large language models.
\newblock \emph{arXiv preprint arXiv:2309.12284}.

\bibitem[{Yue et~al.(2023)Yue, Qu, Zhang, Fu, Huang, Sun, Su, and Chen}]{yue2023mammoth}
Xiang Yue, Xingwei Qu, Ge~Zhang, Yao Fu, Wenhao Huang, Huan Sun, Yu~Su, and Wenhu Chen. 2023.
\newblock Mammoth: Building math generalist models through hybrid instruction tuning.
\newblock \emph{arXiv preprint arXiv:2309.05653}.

\bibitem[{Zhang et~al.(2024)Zhang, Li, and Fan}]{zhang2024mario}
Boning Zhang, Chengxi Li, and Kai Fan. 2024.
\newblock Mario eval: Evaluate your math llm with your math llm--a mathematical dataset evaluation toolkit.
\newblock \emph{arXiv preprint arXiv:2404.13925}.

\bibitem[{Zheng et~al.(2024)Zheng, Zhang, Zhang, Ye, and Luo}]{zheng2024llamafactory}
Yaowei Zheng, Richong Zhang, Junhao Zhang, Yanhan Ye, and Zheyan Luo. 2024.
\newblock Llamafactory: Unified efficient fine-tuning of 100+ language models.
\newblock \emph{arXiv preprint arXiv:2403.13372}.

\bibitem[{Zhou et~al.(2022)Zhou, Sch{\"a}rli, Hou, Wei, Scales, Wang, Schuurmans, Cui, Bousquet, Le et~al.}]{zhou2022least}
Denny Zhou, Nathanael Sch{\"a}rli, Le~Hou, Jason Wei, Nathan Scales, Xuezhi Wang, Dale Schuurmans, Claire Cui, Olivier Bousquet, Quoc Le, et~al. 2022.
\newblock Least-to-most prompting enables complex reasoning in large language models.
\newblock \emph{arXiv preprint arXiv:2205.10625}.

\end{thebibliography}

\clearpage
\onecolumn
\appendix
\section*{Appendix}

\startcontents[sections]
\printcontents[sections]{l}{1}{\setcounter{tocdepth}{2}}

\twocolumn
\section{Future Work} 
MCoT is a framework that empowers LLMs to more efficiently and intelligently in multi-step reasoning. MCoT framework can be applied to complex reasoning tasks, such as long context reasoning~\citep{xiong2023effective,caciularu2023peek, chen2023longlora}. 
MCoT is capable of effectively reducing context information, thereby providing a practical and feasible approach for long context reasoning. This method filters and concentrates historical information, significantly improving the efficiency of processing and analyzing long context. Moreover, when integrated with the Retrieval-Augmented Generation (RAG)~\citep{lewis2020retrieval, gao2023retrieval} technology, MCoT holds significant potential in understanding long context.

\section{Real-world Applications}

Inspired by the strict adherence to the Markov property in formal languages like \texttt{lean}, this work presents significant potential for real-world applications in formal reasoning processes. Our method aligns seamlessly with the detailed, step-by-step nature of formal proofs, such as the tactics used in \texttt{lean}. By generating natural language annotations for each step, our approach enhances the interpretability of complex formal arguments, showcasing its relevance to practical scenarios where both precision and human comprehension are paramount. In particular, this method could be applied in fields such as automated theorem proving, legal reasoning, and software verification, where clear, interpretable reasoning steps are crucial for both validation and decision-making.


\section{The reasoning process of Multi-step reasoning and MCoT}
\label{appendix:comp}

\begin{figure*}[t]
\centering
\includegraphics[width=1.0\textwidth]{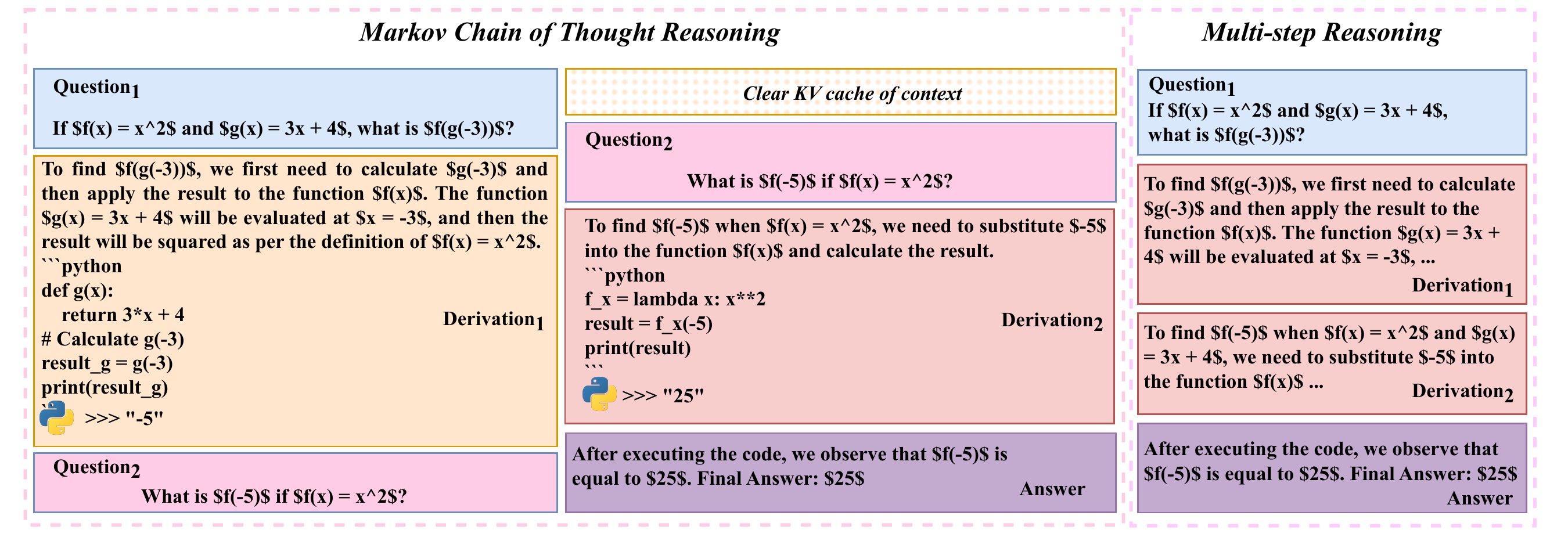}
\caption{\textbf{The reasoning process of two reasoning approachs for mathematical reasoning.} }
\label{fig:approach_example}
\vspace{-4mm}
\end{figure*}

Figure~\ref{fig:approach_example} illustrates the reasoning process using the Markov chain of thought and multi-step reasoning approaches. In multi-step reasoning, the KV cache is retained throughout, causing time and memory demands to increase linearly or quadratically with the number of steps. In contrast, MCoT clears the KV cache after each step, enabling longer chains of thought without a corresponding rise in resource usage. This makes MCoT more efficient, as its time and memory requirements do not scale with the number of reasoning steps.


\section{\texttt{MCoTInstruct} Dataset}
\label{appendix:data}

\subsection{Overview}
Table~\ref{tab:data_compare} presents the detail information of different mathematical reasoning datasets. We list recently popular mathematical reasoning datasets. The \textbf{\#Annotation Num} refers to the number of samples annotated during the dataset construction process. The \textbf{\#Annotation Type} represents three categories: {solution, trajectory, question}. ``solution'' denotes the annotation of a solution, ``trajectory'' refers to the multi-step reasoning path, and ``question'' signifies the annotation of the reduced question. 

Unlike previous works~\cite{yue2023mammoth,gou2023tora,wang2023mathcoder,liao2024mario}, 
our approach requires only the use of GPT-4 for generating a new question, instead of crafting the complete solution that includes text analysis and code snippets. The number of questions that need to be generated is 29$k$.


\subsection{Data Format}
\label{appendix:data_format}

\begin{figure*}[htbp]
\centering
\includegraphics[width=1.0\textwidth]{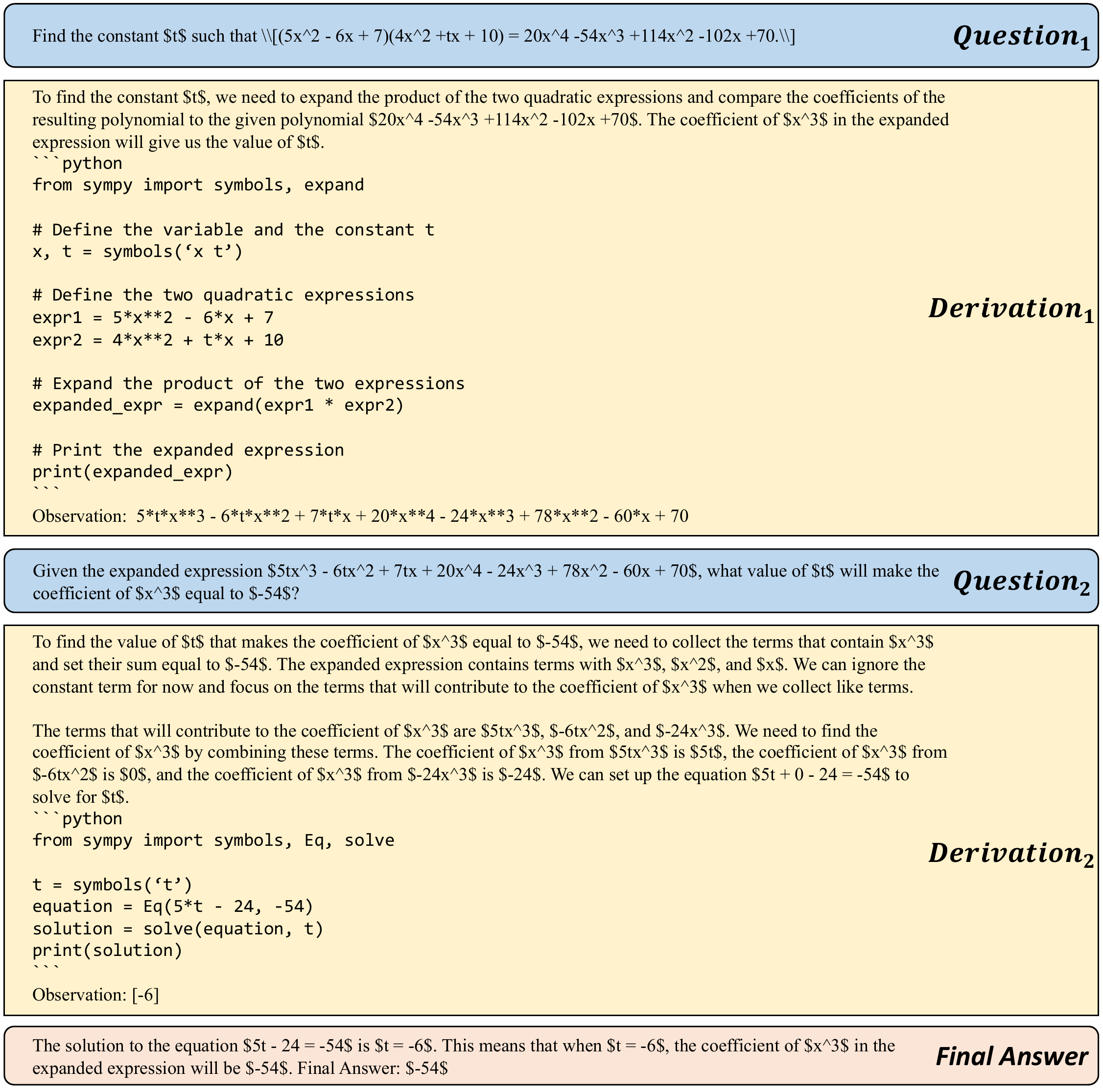}
\caption{The format of Markov Chain of Thought reasoning.}
\label{fig:data_format}
\end{figure*}

Figure~\ref{fig:data_format} displays the format of Markov Chain of Thought reasoning. We decompose the Markov chain into tuples of $(\mathbf{q}_{1}, \mathbf{s}_{1}, \mathbf{q}_{2})$ and $(\mathbf{q}_{2}, \mathbf{s}_{2}, \mathbf{a})$ to form the \texttt{MCoTInstruct} dataset.

\section{Experimental Details}
\subsection{Test Dataset}
\label{appendix:test_data}

We report the information of four test datasets in Table~\ref{tab:datasets_details}. Notably, the four datasets have obvious differences in difficulty and question types and MATH dataset has diverse categories, thereby ensuring the richness and diversity of the testsets. OCWCourses is a set of 272 STEM problems designed for college students, with most questions needing a few steps to solve. 
The GaoKao2023-Math-En dataset includes 385 math problems drawn from the 2023 Chinese National College Entrance Examination, the 2023 American Mathematics Competitions, and the 2023 American College Testing. 
These two OOD datasets are even more tricky to solve than the MATH dataset. 

\begin{figure*}[htbp]
\centering
\includegraphics[width=1.0\textwidth]{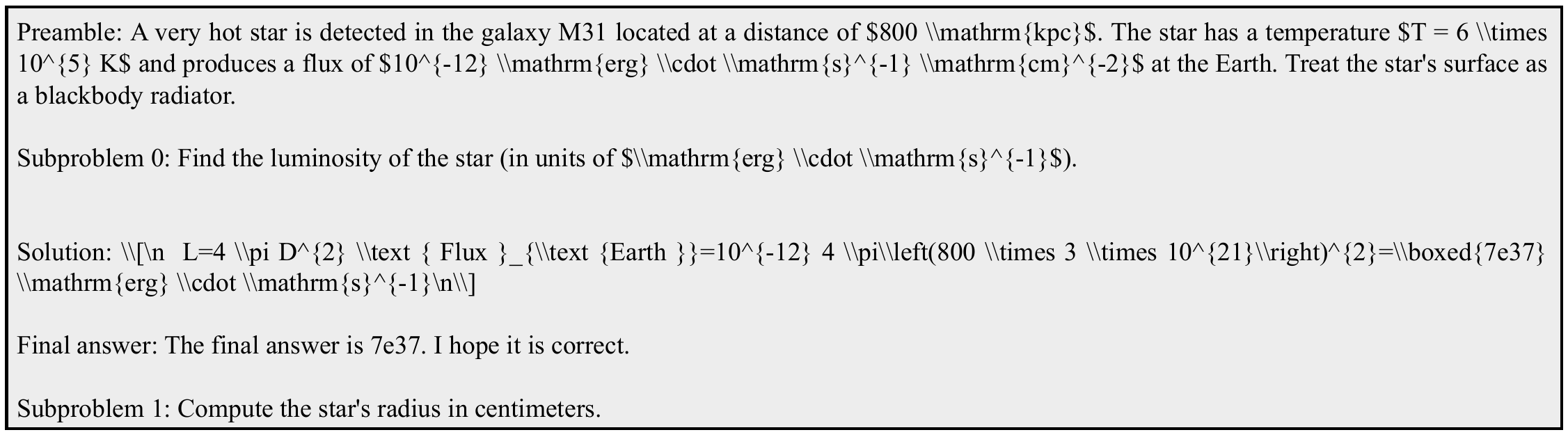}
\caption{The data format of OCWCouses dataset.}
\label{fig:ocw_data_format}
\end{figure*}

Notably, OCW dataset has a somewhat unique question format, which includes a premise, alongside sub problem and their respective solutions, as shown in Figure~\ref{fig:ocw_data_format}.

\begin{table}[!ht]
\setlength{\tabcolsep}{2pt}
\centering
\footnotesize
\renewcommand\arraystretch{1.2}
\begin{tabular}{lcccc}
\toprule[1.2pt]  
\textbf{Dataset} & \textbf{\#Training} & \textbf{\#Test} & \textbf{Category} & \textbf{Domain}  \\
\midrule[0.8pt]
GSM8K & 7473 &  1319 & \xmark & In-domain  \\
MATH & 7500 &  5000 & \cmark & In-domain   \\
OCW & - & 272 & \xmark & Out-of-domain  \\
GaoKao2023 & - &  385 & \xmark & Out-of-domain  \\
\bottomrule[1.2pt]
\end{tabular} 
\caption{\label{tab:datasets_details} The details of four datasets.}
\vspace{-4mm}
\end{table}

\subsection{Baselines}
\label{appendix:baselines}
This comparison included notable models such as OpenAI's GPT-4~\citep{achiam2023gpt} and ChatGPT, Google’s PaLM-2~\citep{anil2023palm}, along with Llama3, Llama2~\citep{touvron2023llama2}, Llemma~\citep{azerbayev2023llemma}, and CodeLlama~\citep{roziere2023code}. 
To establish a fundamental reasoning method baseline, we initially considere Chain of Thought (CoT) prompts~\citep{wei2022chain}. 
Additionally, given our methodology's reliance on the Python code interpreter, we also evaluate the Program of Thought (PoT)~\cite{chen2022program} and Program-aided Language (PAL) model~\citep{gao2023pal}.

For supervised fine-tuning (SFT) models, we categorize them into \textit{single-step reasoning}: Mammoth~\citep{yue2023mammoth}, DeepSeekMath-Instruct~\citep{shao2024deepseekmath} and \textit{multi-step reasoning}: MathCoder~\citep{wang2023mathcoder}, ToRA~\citep{gou2023tora}, MARIO~\citep{liao2024mario} and MathGenie~\citep{lu2024mathgenie}.

\subsection{Fine-tuning Details}
\label{appendix:param_details}
In this work, we finetune all models using the LLaMA-Factory~\citep{zheng2024llamafactory} repository. During this optimization phase, we set the global batch size at 512, the learning rate at 2e-5, and used a cosine learning rate scheduler that included a warm-up phase constituting 3\% of the total training duration, spread over 3 epochs. All models are optimized employing AdamW~\citep{kingma2014adam}. Training for all models was launched with the accelerate~\cite{accelerate} in DeepSpeed ZeRO Stage2~\citep{rajbhandari2021zero} and Flash-Attention 2~\citep{dao2023flashattention} mechanism. The 7B/8B and 70B models are fine-tuned on 8 and 32 NVIDIA A100 80GB GPUs, respectively.

\section{Additional Results and Analyses}
\subsection{The Impact of Self-distillation}
\label{appendix:self-distillation}
Table~\ref{tab:self_distillation} illustrates the impact of self-distillation. It can be observed that self-distillation enhances the coverage of the training sets and elevates the accuracy of the test sets. Specifically, the application of self-distillation led to a 5.5\% increase in data coverage on the MATH training dataset. Furthermore, it achieved a notable 2.5\% improvement in accuracy on the MATH test set and a 0.83\% boost in accuracy on the GSM8K test set.

\setlength{\tabcolsep}{4pt}
\begin{table}[htbp]
\centering
\footnotesize
\renewcommand\arraystretch{1.2}
\begin{center}
    \begin{tabular}{l|cccc}
        \toprule[1.2pt]  
        \textbf{Methods}   & \multicolumn{2}{c}{Trainset Coverage} & \multicolumn{2}{c}{Testset Accuracy}\\
        \midrule[0.8pt]
        Self-Distillation & GSM8K & MATH & GSM8K & MATH \\
        \midrule[0.8pt]
          
        w/o$^\dagger$ & 99.75\% & 77.97\% & 77.94\% & 53.32\%\\
        w/$^\ddagger$ & \textbf{99.85\%} & \textbf{83.46\%} & \textbf{78.77\%} & \textbf{55.78\%} \\
        \bottomrule[1.2pt]
    \end{tabular}
\end{center}
\caption{\label{tab:self_distillation} The impact of Self-distillation on trainset coverage and testset accuracy. Trainset coverage indicates the proportion of the dataset that encompasses questions from the original training sets of GSM8K and MATH, respectively. $^\dagger$ denotes the stage in which self-distillation is not employed, signifying the utilization of seed data. $^\ddagger$ represents the stage that employs self-distillation.}
\end{table}






\subsection{The Analysis on Problem-Solving}
\label{appendix:analysis}
\begin{figure*}[htbp]
\centering
\begin{subfigure}[b]{0.48\linewidth}
    \includegraphics[width=\linewidth]{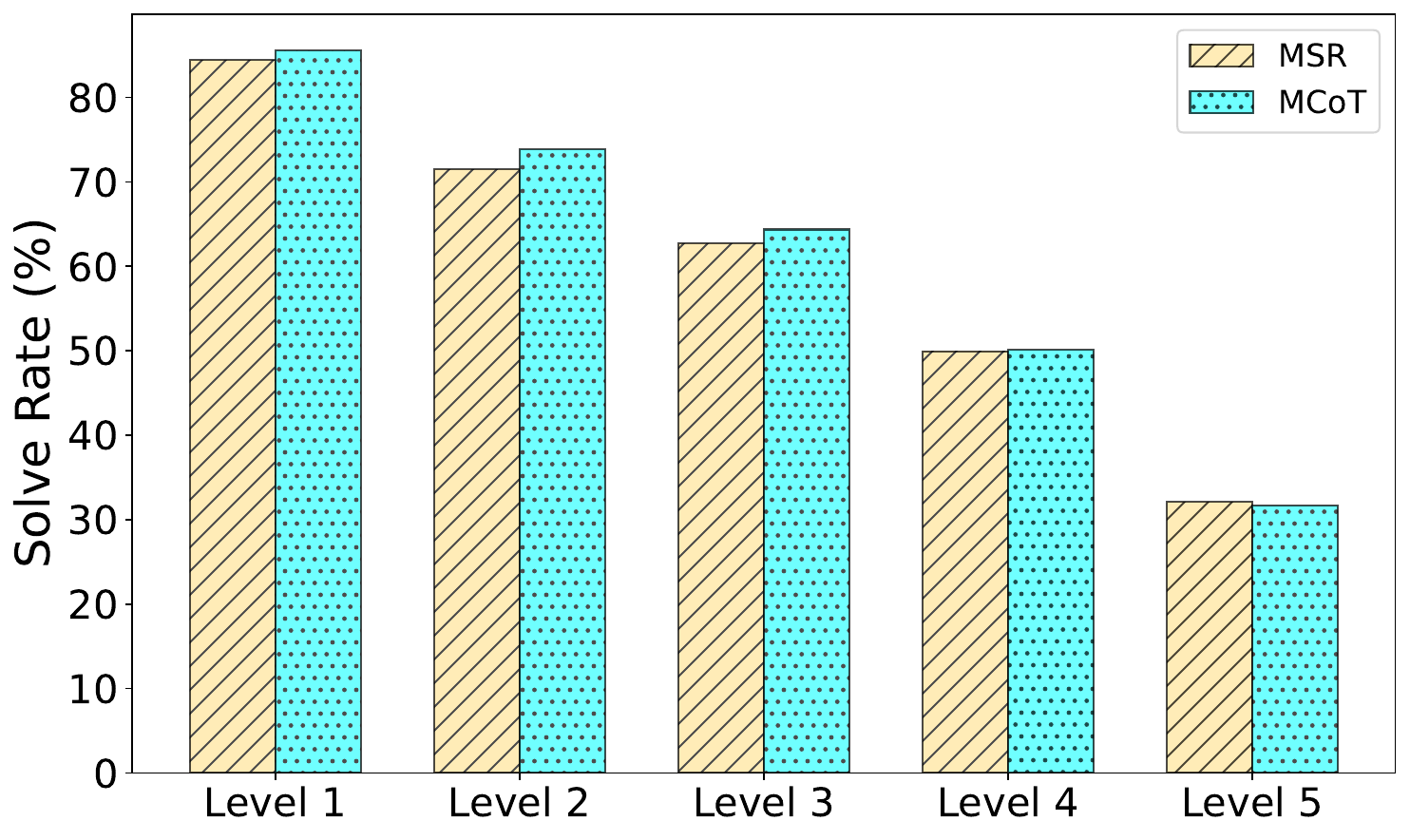}
    \caption{Difficulty Level}
    \label{fig:exp_deepseek_math_level}
\end{subfigure}
\hfill
\begin{subfigure}[b]{0.48\linewidth} 
    \includegraphics[width=\linewidth]{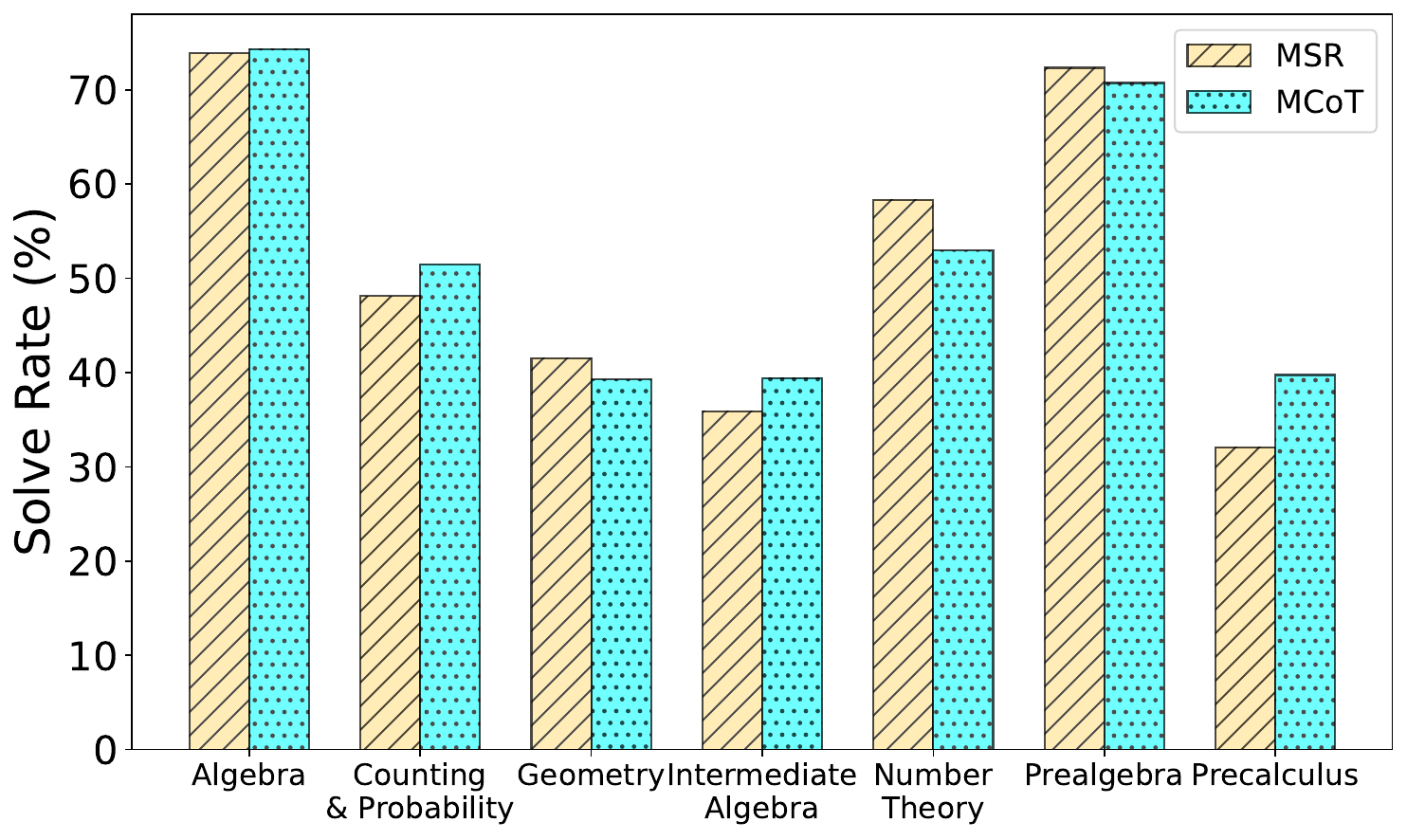} 
    \caption{Different Subject}
    \label{fig:exp_deepseek_math_subject}
\end{subfigure}
\caption{Comparison of problem-solving between MCoT and MSR on MATH test set, with DeepSeekMath-Base\textsubscript{7B} as base model.}
\label{fig:exp_deepseek_math}
\end{figure*}

\begin{figure*}[htbp]
\centering
\begin{subfigure}[b]{0.48\linewidth}
    \includegraphics[width=\linewidth]{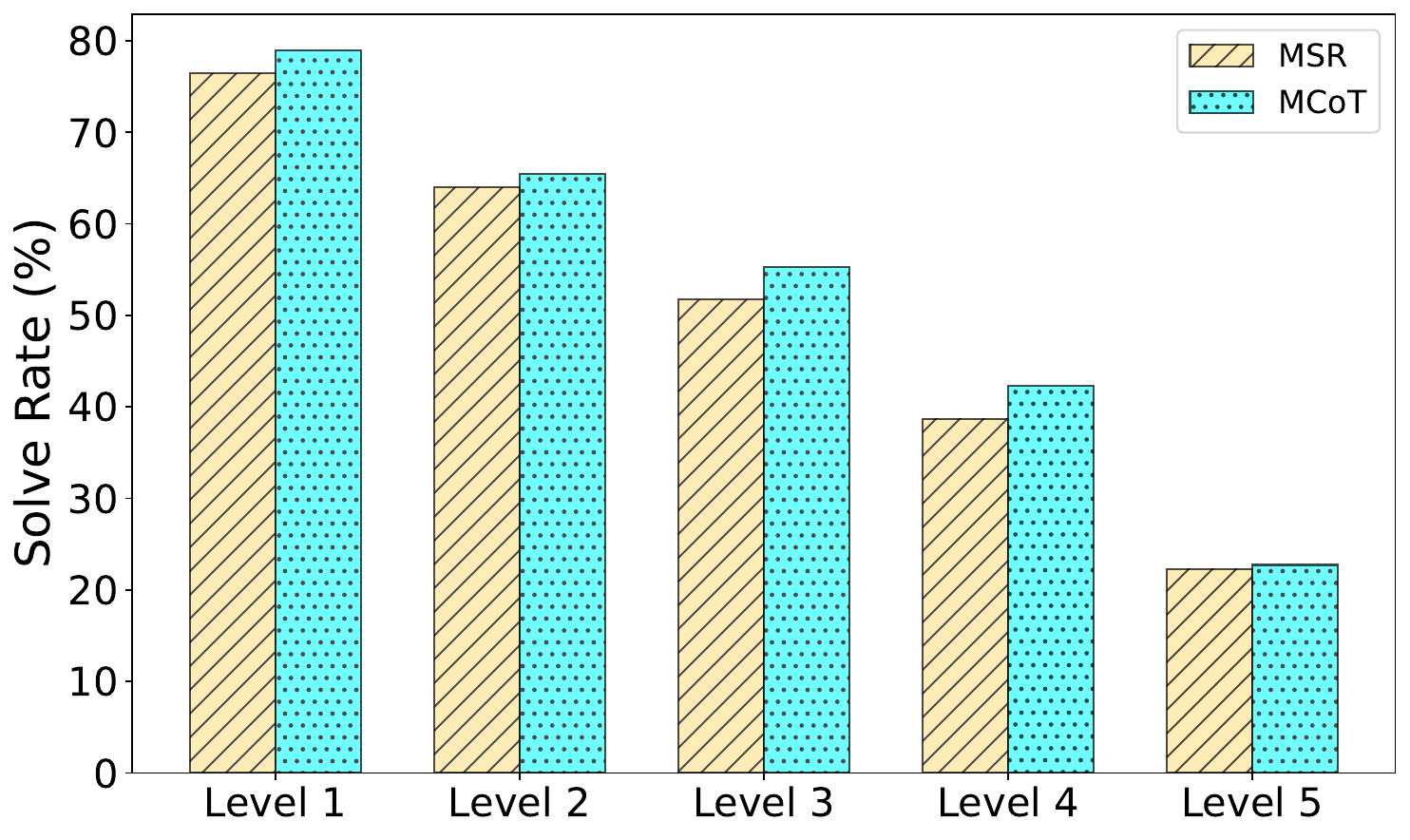}
    \caption{Difficulty Level}
    \label{fig:exp_llama3_math_level}
\end{subfigure}
\hfill
\begin{subfigure}[b]{0.48\linewidth} 
    \includegraphics[width=\linewidth]{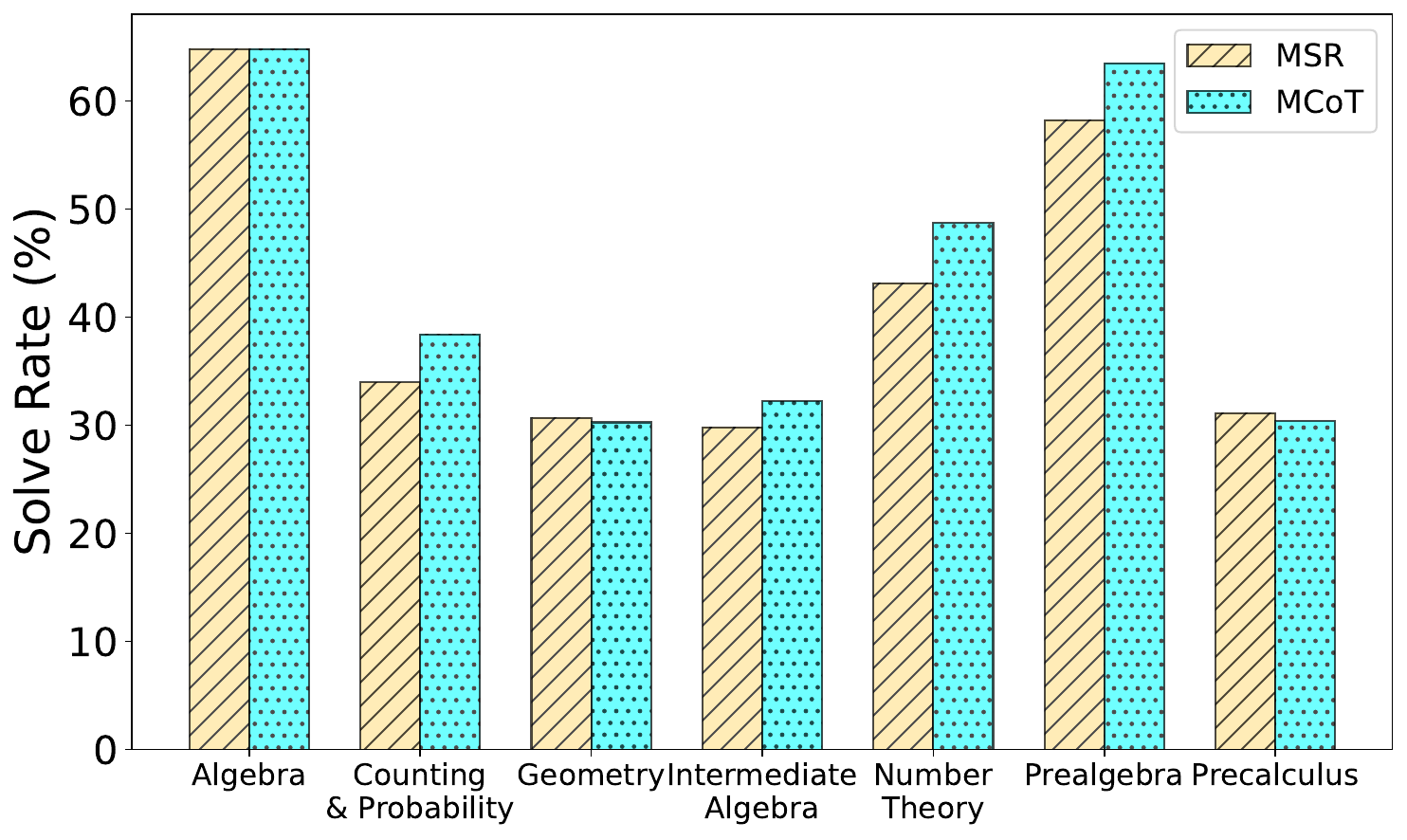} 
    \caption{Different Subject}
    \label{fig:exp_llama3_math_subject}
\end{subfigure}
\caption{Comparison of problem-solving between MCoT and MSR on MATH test set, with Llama3\textsubscript{7B} as base model.}
\label{fig:exp_llama3_math}
\end{figure*}

Figure~\ref{fig:exp_deepseek_math} and Figure~\ref{fig:exp_llama3_math} illustrate the problem-sloving capabilities of MCoT-DeepSeek and MCoT-Llama3, respectively. In terms of difficulty levels as shown in~\ref{fig:exp_deepseek_math_level} and~\ref{fig:exp_llama3_math_level}, MCoT outperforms MSR on levels 1 through 4 and is nearly on par with MSR at level 5. In terms of different subjects as shown in~\ref{fig:exp_deepseek_math_subject} and~\ref{fig:exp_llama3_math_subject}, We discovered that, when using DeepSeek and Llama3 as base models, MCoT consistently outperforms MSR, particularly in the fields of Algebra, Counting \& Probability, and Intermediate Algebra.

\section{Dataset License}
The \texttt{MCoTInstruct} dataset is built on GSM8K~\citep{cobbe2021training} and MATH~\citep{hendrycks2021measuring}. We strictly abide by the corresponding data licenses.

\begin{table*}[htbp]
\setlength{\tabcolsep}{4pt}
\centering
\footnotesize
\vspace{-0.2cm}
\renewcommand\arraystretch{1.2}
\begin{center}
    \begin{tabular}{l|ccccccc}
        \toprule[1.2pt]  
        \textbf{Methods} & \textbf{\#Annotation} & \textbf{\#Annotation} & \textbf{Step} & \textbf{LLM} & \textbf{Source}\\
         & \textbf{Num} & \textbf{Type} &  & \textbf{Used} & \\
        \midrule[0.8pt]
            MetaMATH~\citep{yu2023metamath} & 395$k$ & solution & Single & ChatGPT & MATH \& GSM8K \\
            WizardMath~\cite{luo2023wizardmath} & \textgreater 96$k$ & solution & Single & ChatGPT & MATH \& GSM8K \\
            MAmmoTH~\citep{yue2023mammoth} & 260$k$ & solution  & Single & GPT4 & 14 Datasets with MATH \& GSM8K \\
            MathCoder~\citep{wang2023mathcoder} & 49$k$ & trajectory & Multi & GPT4 & MATH \& GSM8K \\
            ToRA~\citep{gou2023tora} & 16$k$ & trajectory & Multi  & GPT4 & MATH \& GSM8K  \\
            MARIO~\citep{liao2024mario} & 27$k$ & trajectory & Multi  & GPT4 & MATH \& GSM8K \\
            MathGenie~\citep{lu2024mathgenie} & 80$k$ & trajectory & Multi & GPT4 & MATH \& GSM8K \\
            \midrule
            \texttt{MCoTInstruct} & 29$k$ & question & Multi & GPT4 & MATH \& GSM8K \\
        \bottomrule[1.2pt]
    \end{tabular}
\end{center}
\caption{\label{tab:data_compare}Compared with mathematical reasoning datasets.Note that \texttt{MCoTInstruct} dataset instance corresponds to the step level rather than the solution level that may include multiple steps.}
\vspace{-2mm}
\end{table*}

\onecolumn

\section{Case Study}
\label{appendix:case}

To qualitatively analyze responses from Markov Chain of Thought reasoning, we report three cases with \texttt{MCoT-DeepSeek}. In cases 1 and 2, We observed that MCoT does possess some self-correction capabilities. In case 3, during the reduction process, MCoT lost some information, causing the reduced problem to no longer contain all the information of the original problem.

\begin{figure}[!ht]
\centering
\includegraphics[width=0.9\textwidth]{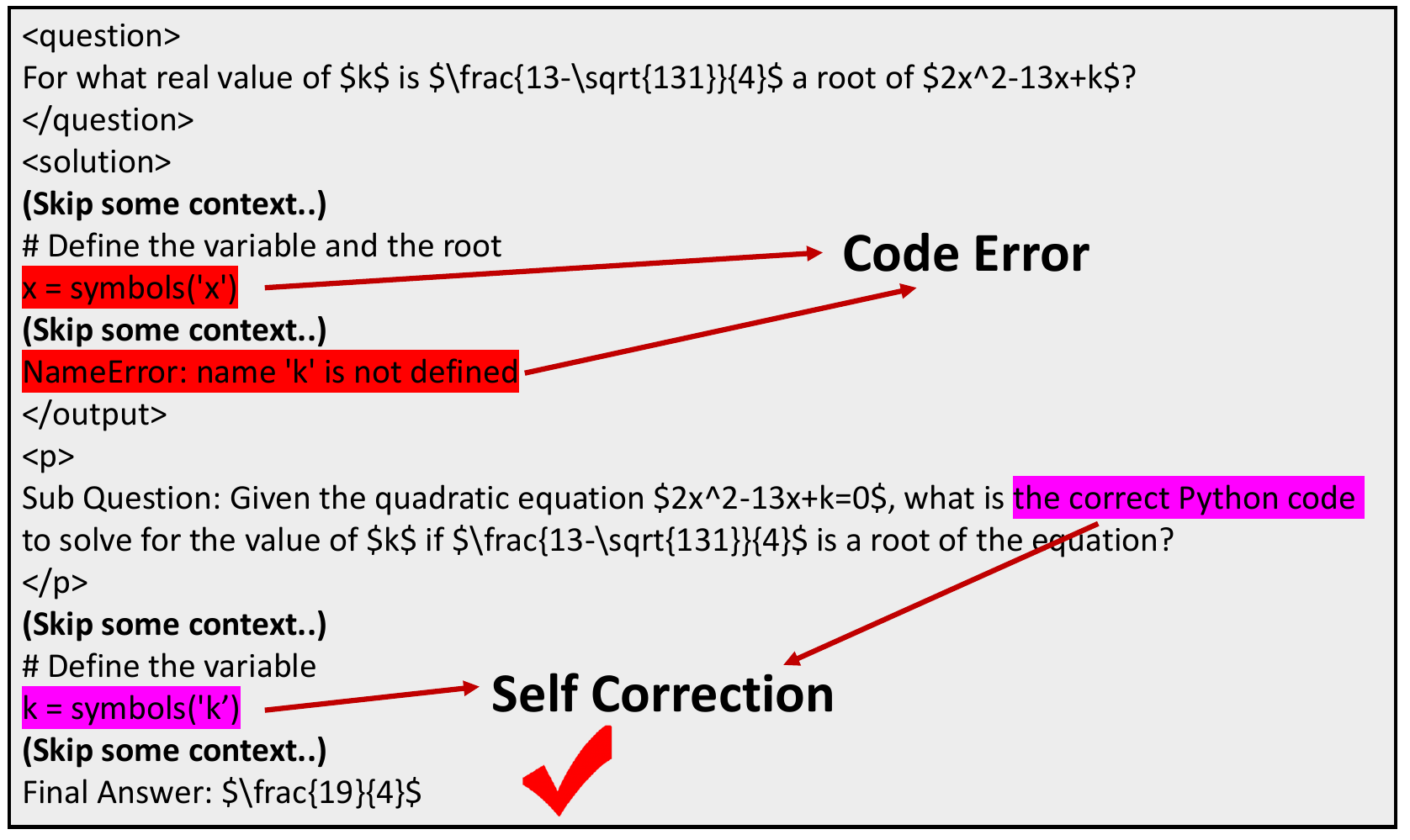}
\caption{The illustration about MCoT's self-correction capability.}
\label{fig:self_correction}
\end{figure}

\subsection{Case 1: Self-correction}
\label{case_1}
\begin{tcolorbox}[colback=lightblue!50!white, colframe=lightblue, title=\textcolor{black}{The Case 1 about MCoT’s self-correcting capability}, width=\textwidth, breakable, label=box_self_correct_case_1]
\footnotesize
\begin{verbatim}
idx: "math_test_74"  level: "Level 3"  type: "Algebra"  standard answer: "\frac{19}{4}"
\end{verbatim}
\textbf{<question>} \newline
For what real value of \(k\) is \(\frac{13-\sqrt{131}}{4}\) a root of \(2x^2-13x+k\)? \newline
\textbf{</question>} \newline

\textbf{<solution>} \newline
To find the value of \(k\) for which \(\frac{13-\sqrt{131}}{4}\) is a root of the quadratic equation \(2x^2-13x+k=0\), we can substitute the root into the equation and solve for \(k\). \newline
\textbf{<code>}
\begin{verbatim}
from sympy import symbols, sqrt, solve

# Define the variable and the root
x = symbols('x')
root = (13 - sqrt(131)) / 4

# Define the quadratic equation
equation = 2*x**2 - 13*x + k

# Substitute the root into the equation and solve for k
k_value = solve(equation.subs(x, root), k)
print(k_value)
\end{verbatim}

\textbf{</code>} \newline

\textbf{<output>} \newline
\hlerror{NameError: name 'k' is not defined} \newline
\textbf{</output>} \newline

Sub Question: Given the quadratic equation \(2x^2-13x+k=0\), what is the correct Python code to solve for the value of \(k\) if \(\frac{13-\sqrt{131}}{4}\) is a root of the equation? \newline
\textbf{</solution>} \newline

\textbf{<solution>} \newline
To find the value of \(k\), we can use the fact that if \(\frac{13-\sqrt{131}}{4}\) is a root of the quadratic equation \(2x^2-13x+k=0\), then the other root must be \(\frac{13+\sqrt{131}}{4}\). This is because the sum of the roots of a quadratic equation is given by \(-\frac{b}{a}\), where \(a\) and \(b\) are the coefficients of \(x^2\) and \(x\) respectively. In our case, \(-\frac{b}{a} = \frac{13}{2}\), so the sum of the roots is \(\frac{13}{2}\). Therefore, we can find \(k\) by multiplying the roots. \newline
\textbf{<code>}
\begin{verbatim}
from sympy import symbols, sqrt, simplify

# Define the variable
k = symbols('k')

# Define the roots
root1 = (13 - sqrt(131)) / 4
root2 = (13 + sqrt(131)) / 4

# Calculate the product of the roots (which gives us k)
product_of_roots = simplify(root1 * root2)
print(product_of_roots)
\end{verbatim}

\textbf{</code>} \newline

\textbf{<output>} \newline
19/8 \newline
\textbf{</output>} \newline

From the result, we can see that the product of the roots is \(\frac{19}{8}\). Since the product of the roots is also equal to \(\frac{k}{a}\), and \(a = 2\) for our quadratic equation, we can find the value of \(k\) by multiplying the product of the roots by \(a\). \newline
Since we have found that the product of the roots is \(\frac{19}{8}\) and \(a = 2\), we can calculate \(k\) by multiplying the product of the roots by \(a\). \newline
Final Answer: \(\frac{19}{4}\) \newline
\textbf{</solution>}
\end{tcolorbox}

\label{case_2}
\subsection{Case 2: Self-correction}
\begin{tcolorbox}[breakable, colback=lightblue!50!white, colframe=lightblue, title=\textcolor{black}{The Case 2 about MCoT’s self-correcting capability}, width=\textwidth, label=box_self_correct_case_2]
\footnotesize
\begin{verbatim}
idx: "math_test_122"  level: "Level 5"  type: "Algebra"  standard answer: "3"
\end{verbatim}
\textbf{<question>} \newline
For how many positive integral values of \(a\) is it true that \(x = 2\) is the only positive integer solution of the system of inequalities: 
\[
\begin{cases}
2x > 3x - 3 \newline
3x - a > -6
\end{cases}
\]
\textbf{</question>} \newline

\textbf{<solution>} \newline
To find the positive integral values of \(a\) for which \(x = 2\) is the only positive integer solution to the system of inequalities, we need to solve the inequalities for \(x\) and then find the constraints on \(a\). \newline
The first inequality \(2x > 3x - 3\) simplifies to \(x < 3\). This tells us that \(x\) must be less than 3. \newline
The second inequality \(3x - a > -6\) can be simplified by adding \(a\) to both sides and then dividing by 3 to get \(x > \frac{a}{3} - 2\). \newline
For \(x = 2\) to be the only positive integer solution, \(x\) must be greater than or equal to 2 but less than 3. This means that \(2 \leq x < 3\). \newline
Therefore, we need to find the range of \(a\) such that \(2 \leq \frac{a}{3} - 2 < 3\). \newline
\textbf{<code>}
\begin{verbatim}
from sympy import symbols, solve, S

# Define the variable
a = symbols('a', real=True, positive=True)

# Inequality 1: 2 <= a/3 - 2
ineq1 = a/3 - 2 >= 2

# Inequality 2: a/3 - 2 < 3
ineq2 = a/3 - 2 < 3

# Solve the inequalities
solution1 = solve(ineq1, a)
solution2 = solve(ineq2, a)

# Find the intersection of the solutions
intersection = (S.Intersection(solution1, solution2)).evalf()
print(intersection)
\end{verbatim}

\textbf{</code>} \newline

\textbf{<output>} \newline
\hlerror{AttributeError: Attribute 'Intersection' was not installed on SymPy registry S} \newline
\textbf{</output>} \newline

Sub Question: Given the inequalities \(2 \leq \frac{a}{3} - 2 < 3\), for how many positive integral values of \(a\) is \(a\) within the range found? \newline
\textbf{</solution>} \newline

\textbf{<solution>} \newline
The given inequalities are \(2 \leq \frac{a}{3} - 2 < 3\). To find the range of \(a\), we need to solve these inequalities for \(a\). \newline
\textbf{<code>}
\begin{verbatim}
from sympy import symbols, solve, S

# Define the variable
a = symbols('a')

# Define the inequalities
ineq1 = 2 <= a/3 - 2
ineq2 = a/3 - 2 < 3

# Solve the inequalities
solution1 = solve(ineq1, a)
solution2 = solve(ineq2, a)

print(solution1, solution2)
\end{verbatim}

\textbf{</code>} \newline

\textbf{<output>} \newline
(12 <= a) \& (a < oo) (-oo < a) \& (a < 15) \newline
\textbf{</output>} \newline

Sub Question: For how many positive integral values of \(a\) is \(a\) within the range \(12 \leq a < 15\)? \newline
\textbf{</solution>} \newline

\textbf{<solution>} \newline
The task is to determine the number of positive integral values of \(a\) that lie within the range \(12 \leq a < 15\). Since \(a\) must be an integer, we can simply list the integers within this range. \newline
The integers that satisfy \(12 \leq a < 15\) are 12, 13, and 14. There are three such integers, so the answer is 3. \newline
Final Answer: \(3\) \newline
\textbf{</solution>}
\end{tcolorbox}

\subsection{Case 3: Bad case}
\label{case_3}
\begin{tcolorbox}[colback=lightblue!50!white, colframe=lightblue, title=\textcolor{black}{The bad case}, width=\textwidth, breakable, label=box_bad_case]
\footnotesize
\begin{verbatim}
idx: "math_test_1601"  level: "Level 4"   type: "Counting & Probability"  standard answer: "50"
\end{verbatim}

\textbf{<question>} \newline
In how many ways can the letters of the word BANANA be rearranged such that the new word does not begin with a B? \newline
\textbf{</question>} \newline

\textbf{<solution>} \newline
To find the number of ways to rearrange the letters of the word BANANA such that the new word does not begin with a B, we can first find the total number of rearrangements of the word BANANA and then subtract the number of rearrangements that start with a B. \newline

\textbf{<code>} \newline
\begin{verbatim}
from math import factorial

# Total number of letters in BANANA is 6 (with A repeating 3 times)
# So the total number of rearrangements is 6! / (3!)
total_rearrangements = factorial(6) // factorial(3)
print(total_rearrangements)
\end{verbatim}

\textbf{</code>} \newline

\textbf{<output>} \newline
120 \newline
\textbf{</output>} \newline

Sub Question: Given that there are 120 ways to rearrange the letters of the word BANANA, how many of these rearrangements do not start with the letter B? \newline

\textbf{</solution>} \newline

\textbf{<solution>} \newline
To find the number of rearrangements of the word BANANA that do not start with the letter B, we can first find the total number of rearrangements and then subtract the number of rearrangements that do start with the letter B. The total number of rearrangements of a word with repeated letters can be calculated using the formula for permutations of a multiset, which is 

\[
\frac{n!}{n_1! \cdot n_2! \cdot ... \cdot n_k!}
\]

where \(n\) is the total number of letters, and \(n_1, n_2, ..., n_k\) are the frequencies of each distinct letter. \newline

\textbf{<code>} \newline
\begin{verbatim}
from math import factorial

# Total number of letters in BANANA
n = 6
# Frequency of each letter: A - 3 times, B - 1 time, N - 2 times
n_A = 3
n_B = 1
n_N = 2

# Total number of rearrangements
total_rearrangements = factorial(n) // (factorial(n_A) * factorial(n_B) * factorial(n_N))
print(total_rearrangements)
\end{verbatim}

\textbf{</code>} \newline

\textbf{<output>} \newline
60 \newline
\textbf{</output>} \newline

\hlerror{Sub Question: If there are 60 ways to rearrange the letters of the word BANANA, how many of these rearrangements start with the letter B?}

\textbf{</solution>} \newline

\textbf{<solution>} \newline
To find the number of rearrangements of the letters of the word BANANA that start with the letter B, we can consider the remaining 5 letters that can be arranged in any order after fixing the first letter B. Since the first letter is fixed as B, we have 5 remaining letters (A, N, A, N, A) to arrange. We need to calculate the number of unique arrangements of these 5 letters. \newline

\textbf{<code>} \newline
\begin{verbatim}
import math

# Calculate the number of unique arrangements of 5 letters
arrangements = math.factorial(5) // (math.factorial(2) * math.factorial(3))
print(arrangements)
\end{verbatim}

\textbf{</code>} \newline

\textbf{<output>} \newline
10 \newline
\textbf{</output>} \newline

From the result, we can see that there are 10 unique arrangements of the 5 remaining letters after fixing the first letter B. \newline

Final Answer: \(10\)\newline
\textbf{</solution>}
\end{tcolorbox}

\section{Prompt Template in GPT-4}
\label{appendix:prompt}
We provide prompts for generating simplified questions using GPT-4 on the MATH dataset and the GSM8K dataset, respectively.
 
\begin{lstlisting}[
    style=casestyle,
    label={lst:math_prompt}, 
    caption={Prompts of reduced question for GPT-4 on MATH dataset.} % 设置标题
]
You will be provided with an original math problem and its intermediate thought, action and observation. Please generate a sub question that should utilize the results of the intermediate thought, action and observation to replace the related information of original problem and delete redundant irrelevant information. The sub question should be wrapped by <subquestion> and </subquestion>.
Here are some examples you may refer to:


The following are 2 demonstration examples.

<original problem> 
Let $f(x)=x^3+3$ and $g(x) = 2x^2 + 2x +1$. What is $g(f(-2))$?
</original problem> 
<intermediate thought> 
To find $g(f(-2))$, we first need to calculate $f(-2)$ and then substitute this result into $g(x)$.
</intermediate thought> 
<intermediate action>
```python
f_x = lambda x: x**3 + 3
result_f = f_x(-2)
result_f
```
</intermediate action> 
<intermediate observation> 
-5
</intermediate observation>
<subquestion> 
what is $g(-5)$ if $g(x) = 2x^2 + 2x + 1$?
</subquestion>

<original problem> 
The perimeter of a rectangular garden is 60 feet. If the length of the field is twice the width, what is the area of the field, in square feet?
</original problem> 
<intermediate thought> 
he perimeter (P) of a rectangle is given by the formula $P = 2l + 2w$, where $l$ is the length and $w$ is the width of the rectangle. We are given that the perimeter is 60 feet and the length is twice the width, which can be expressed as $l = 2w$. We need to find the area (A) of the rectangle, which is given by the formula $A = l \times w$.
</intermediate thought> 
<intermediate action>
```python
from sympy import symbols, Eq, solve

# Define symbols
l, w = symbols('l w')

# Given perimeter
P = 60

# Length is twice the width
length_eq = Eq(l, 2*w)

# Perimeter equation
perimeter_eq = Eq(2*l + 2*w, P)

# Solve the system of equations for l and w
solution = solve((length_eq, perimeter_eq), (l, w))
print(solution)
```
</intermediate action> 
<intermediate observation> 
{l: 20, w: 10}
</intermediate observation>
<subquestion> 
Given that the length of a rectangle is 20 feet and the width is 10 feet, what is the area of the rectangle, in square feet?
</subquestion>

Now! It's your turn.

<original problem> 
A rectangular band formation is a formation with $m$ band members in each of $r$ rows, where $m$ and $r$ are integers. A particular band has less than 100 band members. The director arranges them in a rectangular formation and finds that he has two members left over. If he increases the number of members in each row by 1 and reduces the number of rows by 2, there are exactly enough places in the new formation for each band member. What is the largest number of members the band could have?
</original problem> 
<intermediate thought> 
The problem describes two different rectangular formations. In the first formation, there are $m$ members in each of $r$ rows with 2 members left over. In the second formation, there are $m+1$ members in each of $r-2$ rows, and there are no members left over. This means that the total number of band members can be expressed in two ways:

1. $mr + 2$ (first formation)
2. $(m+1)(r-2)$ (second formation)

Since these two expressions represent the same total number of band members, we can set them equal to each other and solve for one variable in terms of the other. We also know that the total number of band members is less than 100.
</intermediate thought> 
<intermediate action>
```python
from sympy import symbols, Eq, solve

m, r = symbols('m r', integer=True)
eq = Eq(m*r + 2, (m+1)*(r-2))
solution = solve(eq, m)
print(solution)
```
</intermediate action> 
<intermediate observation> 
[r/2 - 2]
</intermediate observation>
\end{lstlisting}

\begin{lstlisting}[
    style=casestyle,
    label={lst:gsm8k_prompt}, 
    caption={Prompts of reduced question for GPT-4 on GSM8K dataset.} % 设置标题
]
You will be provided with an original math problem and its intermediate thought, action and observation. Please generate a sub question that should utilize the results of the intermediate thought, action and observation to replace the related information of original problem and delete redundant irrelevant information. The sub question should be wrapped by <subquestion> and </subquestion>.
Here are some examples you may refer to:


The following are 2 demonstration examples.

<original problem> 
Alexis is applying for a new job and bought a new set of business clothes to wear to the interview. She went to a department store with a budget of $200 and spent $30 on a button-up shirt, $46 on suit pants, $38 on a suit coat, $11 on socks, and $18 on a belt. She also purchased a pair of shoes, but lost the receipt for them. She has $16 left from her budget. How much did Alexis pay for the shoes?
</original problem> 
<intermediate thought> 
To find out how much Alexis paid for the shoes, I need to calculate the total amount spent on the other items she bought and subtract this from her initial budget. The initial budget is $200, and she has $16 left after all her purchases. The equation for this calculation is:

\[ \text{Cost of shoes} = \text{Initial budget} - (\text{Cost of shirt} + \text{Cost of pants} + \text{Cost of coat} + \text{Cost of socks} + \text{Cost of belt} + \text{Money left}) \]
</intermediate thought> 
<intermediate action>
```python
initial_budget = 200
cost_shirt = 30
cost_pants = 46
cost_coat = 38
cost_socks = 11
cost_belt = 18
money_left = 16

other_items = cost_shirt + cost_pants + cost_coat + cost_socks + cost_belt + money_left
print(other_items)
```
</intermediate action> 
<intermediate observation> 
143
</intermediate observation>
<subquestion> 
Alexis had a budget of $200 for her business attire and after purchasing several items, she was left with $16. If the total cost of the other items was $143, how much did she spend on the pair of shoes?
</subquestion>

<original problem> 
James creates a media empire.  He creates a movie for $2000.  Each DVD cost $6 to make.  He sells it for 2.5 times that much.  He sells 500 movies a day for 5 days a week.  How much profit does he make in 20 weeks?
</original problem> 
<intermediate thought> 
To calculate the profit James makes, we need to consider the initial cost of creating the movie, the cost of producing each DVD, the selling price of each DVD, the number of DVDs sold per day, the number of days he sells per week, and the total number of weeks. The profit for each DVD sold is the selling price minus the cost to make it. The total profit is the sum of the profit from each DVD sold minus the initial cost of creating the movie.

The profit per DVD can be calculated as follows:
Profit per DVD = Selling price per DVD - Cost to make per DVD

The total profit can be calculated as follows:
Total profit = (Profit per DVD * Number of DVDs sold per day * Number of days per week * Number of weeks) - Initial cost of creating the movie

Let's calculate the profit per DVD first.
</intermediate thought> 
<intermediate action>
```python
cost_per_dvd = 6
selling_price_per_dvd = 2.5 * cost_per_dvd
profit_per_dvd = selling_price_per_dvd - cost_per_dvd
print(profit_per_dvd)
```
</intermediate action> 
<intermediate observation> 
9.0
</intermediate observation>
<subquestion> 
Given that James makes a profit of $9.0 per DVD, sells 500 DVDs a day for 5 days a week, and the initial cost of creating the movie was $2000, how much total profit does he make in 20 weeks?
</subquestion>

Now! It's your turn.

<original problem> 
James creates a media empire.  He creates a movie for $2000.  Each DVD cost $6 to make.  He sells it for 2.5 times that much.  He sells 500 movies a day for 5 days a week.  How much profit does he make in 20 weeks?
</original problem> 
<intermediate thought> 
To calculate the profit James makes, we need to consider the initial cost of creating the movie, the cost of producing each DVD, the selling price of each DVD, the number of DVDs sold per day, the number of days he sells per week, and the total number of weeks. The profit for each DVD sold is the selling price minus the cost to make it. The total profit is the sum of the profit from each DVD sold minus the initial cost of creating the movie.

The profit per DVD can be calculated as follows:
Profit per DVD = Selling price per DVD - Cost to make per DVD

The total profit can be calculated as follows:
Total profit = (Profit per DVD * Number of DVDs sold per day * Number of days per week * Number of weeks) - Initial cost of creating the movie

Let's calculate the profit per DVD first.
</intermediate thought> 
<intermediate action>
```python
cost_per_dvd = 6
selling_price_per_dvd = 2.5 * cost_per_dvd
profit_per_dvd = selling_price_per_dvd - cost_per_dvd
print(profit_per_dvd)
```
</intermediate action> 
<intermediate observation> 
9.0
</intermediate observation>
\end{lstlisting}

\end{document}